\documentclass{article}
\usepackage[T1]{fontenc}
\usepackage{microtype,graphicx,booktabs,multirow,tabularx,placeins}
\usepackage[table]{xcolor}
\usepackage{amsmath}
\usepackage{amssymb}
\usepackage{mathtools}
\usepackage{amsthm}
\usepackage{xurl}
\usepackage{hyperref}

\usepackage[preprint]{icml2027}

\usepackage[capitalize,noabbrev]{cleveref}

\theoremstyle{plain}

\theoremstyle{definition}

\theoremstyle{remark}

\usepackage[textsize=tiny]{todonotes}

\icmltitlerunning{Zero-Shot Text Detection via Context Perturbation Discrepancy}

\begin{document}
\twocolumn[
  \icmltitle{Beyond Fixed-Context Likelihood: Zero-Shot AI-Generated Text Detection via Context Perturbation Discrepancy}

  \begin{icmlauthorlist}
    \icmlauthor{Yifan Tang}{sch1}
    \icmlauthor{Jingran Li}{sch1}
    \icmlauthor{Yihao Wang}{comp1}
    \icmlauthor{Zhen Yang}{sch1}
    \icmlauthor{Jianyi Liu}{sch1}
    \icmlauthor{Gongshen Liu}{sch2}
    \icmlauthor{Ru Zhang}{sch1}

  \end{icmlauthorlist}

  \icmlaffiliation{comp1}{Beijing Academy of Artificial Intelligence, Beijing, China}
  \icmlaffiliation{sch1}{School of Cyberspace Security, Beijing University of Posts and Telecommunications, Beijing, China}
  \icmlaffiliation{sch2}{School of Cyberspace Security, Shanghai Jiao Tong University, Shanghai, China}

  \icmlcorrespondingauthor{Ru Zhang}{zhangru@bupt.edu.cn}

  \icmlkeywords{AI-generated text detection, zero-shot detection, language models, text provenance}
  \vskip 0.3in
]



\begin{NoHyper} 
  \printAffiliationsAndNotice{}  
\end{NoHyper}

\begin{abstract}
The same increase in token log probability from earlier context can be typical at one position and unusually large at another. 
We introduce Context Perturbation Discrepancy (CPD), a zero-shot detector that measures this gain relative to what a single pretrained language model expects. 
Keeping the passage fixed, CPD scores each observed token under its full prefix and under a short window of recent tokens. 
Under the full-prefix distribution, the expected gain equals the KL divergence from that distribution to the restricted-prefix distribution. 
CPD subtracts this expectation from each observed gain and scales the sum of the signed deviations by its analytical reference standard deviation. 
On unedited human and AI texts from GEN, DetectRL-X, and HART, subtracting the expected gain raises AUROC by 32.78--35.81 percentage points at the same variance scale. 
CPD outperforms seven zero-shot baselines, averaging 94.80\% AUROC and 79.60\% recall at a 5\% false-positive rate.
On original and edited HART passages, the same scores separate texts more clearly by draft origin than by the benchmark's final-expression labels (89.15\% versus 65.50\% AUROC).
Code is available at \url{https://github.com/BUPT-tyf/CPD}.
\end{abstract}

\section{Introduction}
\label{sec:introduction}

Large language models (LLMs) enter the writing process as both drafters and editors \citep{survey}. People revise model-generated drafts, and models polish human-written drafts \citep{gen,hart}. We use \emph{AI-generated} for direct model output and \emph{draft origin} for the human or AI source of a draft before revision. We study two zero-shot tasks with a pretrained language model and no task-specific classifier training. The first distinguishes human writing from direct model output \citep{detectgpt}. The second asks whether the original draft was human- or AI-written, using only the revised passage.

\begin{figure}[t]
    \centering
    \includegraphics[width=\columnwidth]{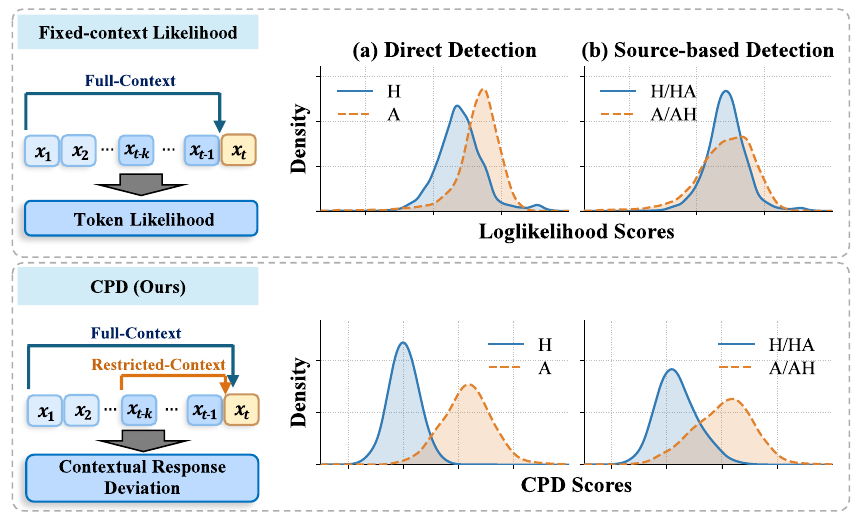}
    \caption{
    Fixed-context likelihood (top) and CPD (bottom).
    Left: token likelihood versus contextual response deviation.
    Right: document score distributions on HART for (a) direct detection (H vs.\ A) and (b) detection by draft origin under editing (H/HA vs.\ A/AH).
    H and A denote original human and AI texts. HA denotes AI-edited human text, and AH denotes AI text humanized by a model, tool, or person.
    The blue curves show the distributions for human-origin texts and the orange curves those for AI-origin texts.
    }
    \label{fig:prob_CR}
\end{figure}

Existing zero-shot detectors use token likelihood and rank, probability curvature, and comparisons between predictive distributions \citep{detectllm,detectgpt,fastdetectgpt,binoculars,nts}. Others examine temperature sensitivity, likelihood at multiple context scales, or dependencies within the text \citep{nts,lar,tcd}. Editing can change the statistical patterns these detectors measure \citep{krishna2023paraphrasing,raid,mage,detectrl}. On HART \citep{hart}, full-prefix likelihood partly separates unedited human and AI texts. Once edited passages are included, the likelihood distributions for the two draft origins overlap substantially (Figure~\ref{fig:prob_CR}, top row). We ask whether the change in a token's probability when earlier context is restored carries evidence of draft origin, and how much of this evidence survives editing.

Consider two tokens with the same probability under their respective full prefixes. With only the most recent tokens available, the first retains that probability and the second becomes much less likely. Restoring earlier context therefore gives the second token a larger gain, even though the full-prefix probabilities are identical. We measure this gain by scoring each observed token twice with the same model. The difference between its full- and restricted-prefix log probabilities is its \emph{contextual response}.

The same gain can exceed the scorer's expectation at one position and fall below it at another. To obtain the \emph{model-expected contextual response} at each position, we average the gains of possible next tokens under the full-prefix distribution. This average equals the KL divergence from the full-prefix distribution to the restricted one. Subtracting it from the observed gain gives a signed \emph{contextual response deviation}. A positive gain yields a negative deviation whenever the scorer expects a larger gain at that position.

Our detector, \textbf{Context Perturbation Discrepancy (CPD)}, sums these signed deviations across the document and divides the sum by its reference standard deviation. Only the prefix visible to the scorer changes, so a single pretrained model can score the passage without generating alternatives. On HART, this comparison separates the two draft origins more clearly than full-prefix likelihood both before and after edited passages are included (Figure~\ref{fig:prob_CR}, lower row).

Our analysis and experiments make three contributions.

(1) \textbf{A prefix-specific reference for context gain.} We derive the gain a scorer expects when earlier context is restored. Subtracting this KL reference gives the observed token a signed deviation whose expectation under the scorer is zero at every prefix. The document score sums the signed deviations over the scored positions, so gains above expectation at some prefixes can offset shortfalls elsewhere in the passage.

(2) \textbf{What each context contributes.} On all three direct-detection benchmarks, centering improves document ranking even when we simply average token scores. We decompose the centered gain into two log-likelihood residuals and show that their difference gives higher AUROC than either alone. For the short prefix, a random earlier token outperforms the immediate predecessor when the restricted-context residual is scored alone. On all three benchmarks, the predecessor yields higher AUROC when each short-prefix residual is paired with the same full-context residual.

(3) \textbf{Draft-origin detection before and after revision.} CPD outperforms seven zero-shot baselines on unedited texts from GEN, DetectRL-X, and HART. Regrouping the same HART passages shows that its scores track draft origin more closely than the benchmark's final-expression labels. In matched revisions, heavier AI editing of human drafts on GEN progressively weakens separation from AI output, while expanding AI drafts on DetectRL-X raises recall at a 5\% false-positive rate and polishing them lowers it.

\section{Related Work}
\label{sec:related_work}

\noindent\textbf{Statistical and contextual detection of AI-generated text.}
Statistical detectors use token likelihood and rank \citep{gltr,detectllm}, sensitivity to decoding settings \citep{temptest,nts}, and variation in token statistics across positions and scales \citep{lastde,temporaltomography}. \citet{simpson} calibrate token scores within regions of a model's hidden space. On a shared prefix, Binoculars compares observed perplexity with cross-perplexity between two models \citep{binoculars}. \citet{chen2026dualnetwork} compare log likelihoods across preference alignments, while \citet{liu2025irm} interpret the instruction-tuned model's log-likelihood ratio to its base model as an implicit reward. Alignment Imprint weights the aligned-to-base log-likelihood ratio by token self-information and standardizes it using the base model's conditional moments \citep{wu2026alignmentimprint}.

LAR learns to combine token likelihood features across nested prefixes \citep{lar}, while TCD uses attention entropy to weight token scores \citep{tcd}. Fast-DetectGPT centers observed log likelihood under a conditional sampling distribution \citep{fastdetectgpt}. \citet{finite_sample_tests} compare log-perplexity with mean predictive entropy to test a candidate generator, deriving finite-sample error bounds under assumptions on token probabilities and source separation. CPD centers the log-probability gain an observed token receives when earlier context is restored to the same scoring model. Its reference is the gain expected under the full-prefix distribution at that position.

\noindent\textbf{Perturbation-based detection.}
DetectGPT estimates probability curvature by scoring perturbed versions of a passage \citep{detectgpt}. Other methods compare the passage with a syntactically reordered version \citep{textreorder} or with a revision produced by ChatGPT \citep{zhu2023beat}. \citet{learning2rewrite} train a rewriter to amplify differences in edit distance between human and AI texts. TOCSIN measures the semantic change after random token deletion and uses it to rescale a detector's score \citep{tocsin}. D\&R reconstructs a passage in one black-box model call after shuffling tokens within chunks \citep{d_and_r}. Its classifier uses semantic similarity and word-order agreement between the reconstruction and the original. CPD changes only the prefix visible to the scorer. Both predictions concern the same observed token, so scoring requires no revision or reconstruction of the original passage.

\noindent\textbf{Draft-origin detection under editing.}
Paraphrasing and rewriting can change how reliably detectors identify machine-generated passages \citep{krishna2023paraphrasing,raid,mage,detectrl}. PRDetect uses syntactic structure to detect AI-generated text after perturbation \citep{prdetect}. Editing can also start from a human draft. On GEN, \citet{gen} find that AI editing of human writing leaves a different stylometric trace from direct generation. HART explicitly distinguishes content from expression and proposes detecting AI involvement along both dimensions \citep{hart}. We use HART to ask which distinction a single CPD score captures, grouping the same scored passages first by draft origin and then by final expression. Matched revisions on GEN and DetectRL-X let us follow how the evidence for draft origin changes as human drafts undergo heavier AI editing or the same set of AI drafts undergoes condensation, expansion, or polishing.

\begin{figure*}[t]
    \centering
    \includegraphics[width=0.98\textwidth]{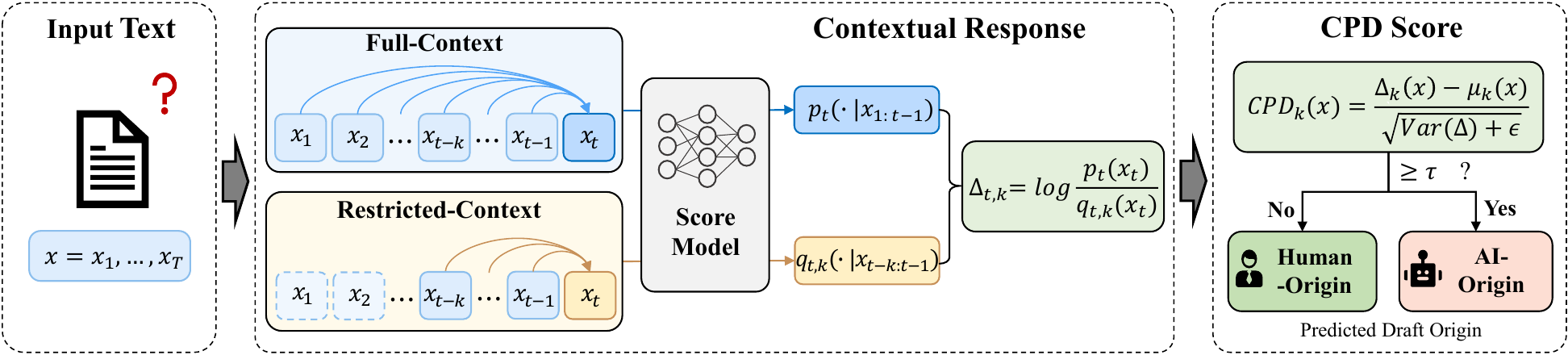}
    \caption{
CPD overview. The same model scores each observed token $x_t$ under its full prefix and the last $k$ tokens of that prefix. At each position, CPD subtracts the expected log-probability gain, given by the KL divergence between the two predictive distributions, from the observed gain. It sums and normalizes these deviations to obtain a document score. $\mathrm{Var}$ denotes the reference variance in Eq.~\eqref{eq:cpd_variance}. The detector assigns AI origin to documents at or above a separately selected threshold and human origin to those with scores below it.
}
    \label{fig:cpd_overview}
\end{figure*}

\section{Methodology}
\label{sec:method}

CPD constructs token-level detection evidence by
centering observed contextual responses against their
model-expected values (Figure~\ref{fig:cpd_overview}).
We first define responses under full and restricted
contexts (Section~\ref{sec:context_perturbation}),
derive their expectations and deviations
(Section~\ref{sec:response_deviation}), and then
aggregate and normalize the deviations into a document
score (Section~\ref{sec:cpd_detection}).

\subsection{Problem Formulation and Context Perturbation}
\label{sec:context_perturbation}

Given an observed text $x=(x_1,\ldots,x_T)$ and a pretrained
autoregressive language model $P_\theta$, we seek a document-level
detection score without task-specific classifier training or access
to the original generator. In direct detection, we distinguish
human-written from directly AI-generated texts. When we evaluate
draft origin after editing, the detector scores the revised text $x$
and the target label identifies who produced the draft before revision.

Models use observations to predict robot actions and future world states~\citep{wang2026vlaadapter,orca2026}.
In autoregressive language modeling, the prefix provides the context for predicting the next token.
CPD measures how earlier tokens affect the probability of the observed token $x_t$ by scoring it
with $P_\theta$ under the full prefix and under only its last $k$ tokens.
\begin{equation}
\label{eq:context_distributions}
\begin{aligned}
p_t(v) &= P_\theta(v\mid x_{1:t-1}),\\
q_{t,k}(v) &= P_\theta(v\mid x_{t-k:t-1}),
\end{aligned}
\end{equation}
where $v\in\mathcal V$ denotes a candidate vocabulary token
and $k$ is the number of retained preceding tokens.
We consider positions $\mathcal I_k=\{k+1,\ldots,T\}$
for $T>k$. At the first scored position, $t=k+1$, both prefixes
contain the same $k$ tokens, so the two next-token distributions are identical.

We define the \emph{contextual response} of a candidate
token $v$ as
\begin{equation}
\label{eq:contextual_response}
\Delta_{t,k}(v)
=
\log p_t(v)-\log q_{t,k}(v)
=
\log\frac{p_t(v)}{q_{t,k}(v)}.
\end{equation}
For the observed token $x_t$, $\Delta_{t,k}(x_t)$ measures the change
in log probability when we restore the context preceding the last $k$ tokens.
A positive response means the earlier context raises the predicted probability
of that token. A negative response means the token is less likely under the full prefix.

\subsection{Contextual Response Deviation}
\label{sec:response_deviation}

At position $t$, we use the full-context distribution $p_t$
as the reference for the next token. Each candidate $v$ gains
$\Delta_{t,k}(v)$ in log probability when earlier context is
restored. Weighting each gain by the full-prefix probability $p_t(v)$ gives the expected log-probability gain for this particular prefix,
\begin{equation}
\label{eq:expected_response}
\begin{aligned}
\mu_{t,k}
&=\mathbb E_{v\sim p_t}[\Delta_{t,k}(v)]\\
&=\sum_{v\in\mathcal V}p_t(v)
\log\frac{p_t(v)}{q_{t,k}(v)}\\
&=D_{\mathrm{KL}}(p_t\Vert q_{t,k}).
\end{aligned}
\end{equation}
This KL divergence gives a single \emph{model-expected contextual
response} for all candidate tokens at that prefix.

We define the \emph{contextual response deviation} by
subtracting this expectation from the observed response:
\begin{equation}
\label{eq:response_deviation}
\begin{aligned}
\delta_{t,k}(x_t)
&=
\Delta_{t,k}(x_t)-\mu_{t,k}\\
&=
\log\frac{p_t(x_t)}{q_{t,k}(x_t)}
-
D_{\mathrm{KL}}(p_t\Vert q_{t,k}).
\end{aligned}
\end{equation}
Under the scorer's own next-token distribution, the deviations
average to zero, $\mathbb E_{v\sim p_t}[\delta_{t,k}(v)]=0$.
Let $r_t$ denote the source's next-token distribution at this prefix.
As shown in Appendix~\ref{app:expectation_centering}, the expected deviation under $r_t$
weights each token's contextual gain by the difference $r_t(v)-p_t(v)$,
\begin{equation}
\label{eq:source_expectation}
\mathbb E_{v\sim r_t}[\delta_{t,k}(v)]
=\sum_{v\in\mathcal V}(r_t(v)-p_t(v))
\log\frac{p_t(v)}{q_{t,k}(v)}.
\end{equation}
A source that shifts probability mass from tokens with smaller
contextual gains to tokens with larger gains, relative to $p_t$,
has positive expected deviation. Centering preserves the ordering
of tokens at a fixed prefix, but the centering term varies across
prefixes. For two passages with the same total observed gain, the
one with the smaller total expected gain has the larger sum of
deviations.

\subsection{Document-Level Aggregation and Normalization}
\label{sec:cpd_detection}

At the document level, we ask whether the observed tokens, taken together,
gain more or less from the full context than the scoring model expects
at their respective positions. We measure this net departure by summing
the signed token-level deviations over $\mathcal I_k$.
Let $\Delta_k(x)=\sum_{t\in\mathcal I_k}\Delta_{t,k}(x_t)$
and
$\mu_k(x)=\sum_{t\in\mathcal I_k}\mu_{t,k}$.
Their difference, $\delta_k(x)=\Delta_k(x)-\mu_k(x)$, equals the sum of the observed token deviations.

To obtain a reference scale for this sum, we hold the observed
prefixes fixed and define hypothetical reference tokens
$\tilde{x}_t\sim p_t$ that are independent across positions
(Appendix~\ref{app:reference_variance}).
Define their accumulated response as
$\Delta_k(\tilde{x};x)=\sum_{t\in\mathcal I_k}\Delta_{t,k}(\tilde{x}_t)$.
The variance of this sum follows from independence of the reference tokens,
\begin{equation}
\label{eq:cpd_variance}
\operatorname{Var}[\Delta_k(\tilde{x};x)]
=
\sum_{t\in\mathcal I_k}
\operatorname{Var}_{v\sim p_t}[\Delta_{t,k}(v)].
\end{equation}
We compute each variance analytically under $p_t$ as the second response moment minus the squared expectation. Subtracting the constant $\mu_{t,k}$ at each fixed prefix leaves the individual response variances and their sum unchanged.

Normalizing the sum gives the document-level \emph{CPD} score
\begin{equation}
\label{eq:cpd}
\boxed{
\operatorname{CPD}_k(x)
=
\frac{
\Delta_k(x)-\mu_k(x)
}{
\sqrt{
\operatorname{Var}[\Delta_k(\tilde{x};x)]
+\epsilon
}
}
}
\end{equation}
Here, $\epsilon>0$ ensures numerical stability.
The score expresses the sum of signed deviations in units of
the reference standard deviation, which reflects both the number
of scored positions and their individual response variances.
Higher CPD scores indicate stronger evidence of AI origin. This uncalibrated score ranks documents directly. For binary classification, we compare each document's score with a
threshold selected for the desired operating point (Figure~\ref{fig:cpd_overview}).

\section{Experiments}
\label{sec:experiments}

We first evaluate CPD on unedited human and AI texts. On HART, we compare how well the same scores distinguish draft origin and final expression. We use matched revisions of human drafts on GEN and AI drafts on DetectRL-X to measure how rewriting affects draft-origin discrimination.

\subsection{Experimental Setup}
\label{sec:exp_setup}

\paragraph{Datasets.}
The English tests compare unedited human texts (H) with direct AI outputs (A) on GEN~\citep{gen}, DetectRL-X~\citep{detectrl}, and HART~\citep{hart}. GEN's prompt-disjoint development and test splits each contain 2,000 texts per class before score filtering. Each test prompt pairs one human text with one model output, yielding 249--251 pairs per generator. The official DetectRL-X and HART tests contain 6,000 and 2,000 texts per class, respectively. HART also includes AI-edited human drafts (HA) and humanized AI drafts (AH). The first letter denotes draft origin, and humanization can involve a model, a tool, or a person (Appendix~\ref{sec:dataset_details}).

\paragraph{Detectors and scoring.}
We score passages without prompts or metadata, using at most 1,024 tokens and BF16 inference. CPD uses Falcon-7B~\citep{falcon} for both prefixes. We score from the second token with $k=1$, which maximized AUROC among $k\in\{1,2,4,8\}$ on GEN development data. Loglikelihood~\citep{gltr}, LRR~\citep{detectllm}, TempTest~\citep{temptest}, and NTS~\citep{nts} share CPD's backbone. Fast-DetectGPT~\citep{fastdetectgpt} and Lastde++~\citep{lastde} use Falcon-7B for the sampling distribution and Falcon-7B-Instruct for scoring. In Binoculars~\citep{binoculars}, the same models serve as observer and performer, respectively. Single-Fast uses Falcon-7B for both Fast-DetectGPT roles in direct detection. Appendices~\ref{app:baseline_settings} and~\ref{app:context_window_results} give detector settings and window-size results.

\paragraph{Evaluation protocol.}
We fix score directions before testing (higher indicates AI origin), including when regrouping HART by final expression. We report the area under the ROC curve (AUROC) and TPR@5\% as percentages. TPR@5\% is the highest test recall at a false-positive rate (FPR) of at most 5\%. For threshold transfer, each detector keeps its GEN development threshold targeting 5\% FPR (Appendix~\ref{app:threshold_transfer}). Macro averages weight benchmarks equally. We compute gains in percentage points from unrounded scores, using the strongest baseline for each metric. Appendix~\ref{sec:data_eval_settings} details sampling and class balancing.

\begin{table*}[t]
\caption{Zero-shot detection of unedited human texts (H) and direct AI outputs (A) on GEN, DetectRL-X, and HART. AUROC and TPR@5\% are percentages. Single-Fast uses the same model for both Fast-DetectGPT roles, and $^\dagger$ marks dual-model configurations. Macro Avg. weights the benchmarks equally. Absolute Gain gives CPD's lead over the best baseline in each column, in percentage points.}
\label{tab:main_results}
\centering
\small
\setlength{\tabcolsep}{2.8pt}
\renewcommand{\arraystretch}{0.95}
\begin{tabular*}{\textwidth}{@{\extracolsep{\fill}}lcccccccc@{}}
\toprule
\multirow{2}{*}{Method}
& \multicolumn{2}{c}{GEN}
& \multicolumn{2}{c}{DetectRL-X}
& \multicolumn{2}{c}{HART}
& \multicolumn{2}{c}{Macro Avg.} \\
\cmidrule(lr){2-3}
\cmidrule(lr){4-5}
\cmidrule(lr){6-7}
\cmidrule(lr){8-9}
& AUROC & TPR@5\%
& AUROC & TPR@5\%
& AUROC & TPR@5\%
& AUROC & TPR@5\% \\
\midrule

\rowcolor{gray!20}
\multicolumn{9}{c}{\textbf{Zero-Shot Detectors}} \\

Loglikelihood~\citep{gltr}
& 43.04 & 8.65
& 55.52 & 8.72
& 75.77 & 15.20
& 58.11 & 10.86 \\

LRR~\citep{detectllm}
& 47.94 & 13.65
& 53.33 & 10.57
& 81.25 & 49.00
& 60.84 & 24.41 \\

TempTest~\citep{temptest}
& 43.91 & 19.90
& 72.74 & 21.80
& 87.69 & 74.85
& 68.12 & 38.85 \\

NTS~\citep{nts}
& 67.75 & 33.55
& 72.73 & 19.92
& 93.30 & 82.15
& 77.92 & 45.21 \\

Single-Fast~\citep{fastdetectgpt}
& 42.79 & 21.30
& 72.72 & 25.30
& 87.71 & 76.60
& 67.74 & 41.07 \\

Binoculars~\citep{binoculars}$^\dagger$
& 46.38 & 23.80
& 68.42 & 9.22
& 91.09 & 85.80
& 68.63 & 39.61 \\

Fast-DetectGPT~\citep{fastdetectgpt}$^\dagger$
& 47.04 & 26.50
& 69.39 & 15.70
& 90.78 & 84.25
& 69.07 & 42.15 \\

Lastde++~\citep{lastde}$^\dagger$
& 68.52 & 43.35
& 68.40 & 11.85
& 93.54 & 85.60
& 76.82 & 46.93 \\

\midrule
\textbf{CPD$_{k=1}$ (Ours)}
& \textbf{94.18} & \textbf{80.25}
& \textbf{91.90} & \textbf{65.60}
& \textbf{98.31} & \textbf{92.95}
& \textbf{94.80} & \textbf{79.60} \\

\textit{Absolute Gain ($\uparrow$)}
& +25.66 & +36.90
& +19.16 & +40.30
& +4.76 & +7.15
& +16.87 & +32.67 \\
\bottomrule
\end{tabular*}
\end{table*}

\subsection{Main Results: Zero-Shot Detection}
\label{sec:main_results}

CPD leads on both metrics across all three benchmarks (Table~\ref{tab:main_results}). It reaches 94.80\% macro AUROC (16.87 points above NTS) and 79.60\% macro TPR@5\% (32.67 points above Lastde++). For each metric and benchmark, CPD's 95\% bootstrap interval lies above those of all baselines (Appendix~\ref{app:bootstrap_ci}). With thresholds calibrated on GEN development data, CPD leads in balanced accuracy on all three tests, with its highest FPR on DetectRL-X (6.95\%, Table~\ref{tab:fixed_threshold}).

At 5\% FPR, CPD's recall ranges from 65.60\% on DetectRL-X (40.30 points above runner-up Single-Fast, its largest lead) to 92.95\% on HART. CPD leads both metrics on seven of eight GEN generators. On GPT-OSS-20B, NTS leads with 60.24\% recall versus CPD's 37.35\% (Table~\ref{tab:gen_generator_results}).
With score directions fixed, five baselines fall below 50\% AUROC on the same five GEN subsets and exceed 50\% on the other three. Falcon-7B log-likelihood typically ranks GPT-OSS-20B outputs below human texts (2.84\% AUROC) and GPT-OSS-120B outputs above them (78.34\%).

With Falcon-7B, CPD outperforms Single-Fast on both metrics across benchmarks. Table~\ref{tab:component_decomposition} isolates the contribution of restricted context. TCD weighting~\citep{tcd} slightly improves CPD's recall on HART and lowers both metrics elsewhere. TOCSIN~\citep{tocsin} improves dual-model Fast-DetectGPT's recall only on DetectRL-X and lowers both CPD metrics on every benchmark, with the largest recall loss on GEN (24.5 points, Table~\ref{tab:tcd_fast_comparison}).

\subsection{Draft Origin under Editing}
\label{sec:Source_hart}

\begin{table*}[t]
\caption{Draft-origin detection and final-expression diagnostics on HART. Each cell reports AUROC and TPR@5\%, both in percent. H denotes unedited human text, A direct AI output, HA an AI-edited human draft, and AH a humanized AI draft. A slash in a column heading pools the indicated categories. In each comparison, the first group supplies the negatives and the second the positives.}
\label{tab:hart_origin}
\centering
\small
\setlength{\tabcolsep}{2.5pt}
\renewcommand{\arraystretch}{0.90}
\begin{tabular*}{\textwidth}{@{\extracolsep{\fill}}lccccc@{}}
\toprule
\multirow{2}{*}{Method}
& \multicolumn{4}{c}{Draft-Origin Detection}
& \multicolumn{1}{c}{Expression Diagnostic} \\
\cmidrule(lr){2-5}
\cmidrule(lr){6-6}
& H/HA vs.\ A/AH
& HA vs. A
& H vs. AH
& HA vs. AH
& H/AH vs.\ A/HA \\
\midrule
LogLikelihood  & 52.12 / 11.48 & 73.44 / 26.50 & 34.58 / 0.55 & 24.67 / 1.00 & 74.19 / 20.13 \\
LRR            & 56.51 / 25.56 & 80.35 / 48.00 & 33.59 / 3.00 & 30.81 / 2.85 & 72.68 / 29.58 \\
TempTest       & 64.47 / 40.19 & 82.84 / 65.25 & 50.19 / 17.36 & 37.13 / 8.40 & 76.25 / 34.88 \\
NTS            & 79.74 / 50.54 & 87.42 / \textbf{66.45} & 76.96 / 41.72 & 61.26 / 19.61 & 70.12 / 23.93 \\
Binoculars     & 70.62 / 43.81 & 84.32 / 60.65 & 62.25 / 39.02 & 44.79 / 9.05 & 76.84 / 33.73 \\
Fast-DetectGPT & 70.44 / 47.31 & 83.79 / 62.80 & 61.89 / 38.87 & 45.27 / 16.86 & 75.95 / 27.83 \\
Lastde++       & 75.12 / 48.01 & 86.84 / 63.75 & 69.01 / 37.72 & 51.06 / 16.76 & 74.74 / 24.98 \\
\midrule
\textbf{CPD$_{k=1}$ (Ours)}   & \textbf{89.15} / \textbf{62.39} & \textbf{90.82} / 65.90 & \textbf{90.80} / \textbf{73.74} & \textbf{76.68} / \textbf{45.32} & 65.50 / 6.88 \\
\bottomrule
\end{tabular*}
\end{table*}

\begin{figure}[t]
    \centering
    \includegraphics[width=\columnwidth]
    {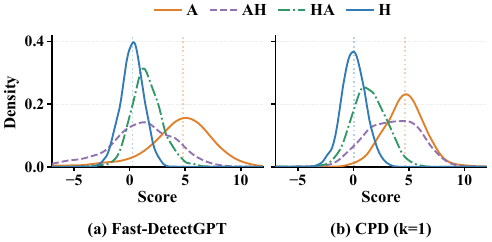}
    \caption{
HART score distributions for (a) Fast-DetectGPT and (b) CPD ($k=1$). H/HA are human-origin and AH/A are AI-origin texts. Higher scores indicate stronger evidence of an AI draft.
}
\label{fig:content-analysis}   
\end{figure}

Changing only the HART labels from draft origin (H/HA versus A/AH) to final expression (H/AH versus A/HA) moves CPD from the highest to the lowest AUROC among the eight detectors (89.15\% versus 65.50\%, Table~\ref{tab:hart_origin}). With FPR capped at 5\% on each ROC curve, recall is 62.39\% for draft origin and 6.88\% for final expression. The score distributions in Figure~\ref{fig:content-analysis} show why draft origin and final expression give different recall values. Under CPD, humanized AI drafts score closer to unedited AI drafts, and AI-edited human drafts score closer to unedited human drafts. Fast-DetectGPT assigns humanized AI drafts scores that overlap substantially with those for human-origin texts.

\paragraph{Discrimination after editing.}
CPD leads in AUROC on all four draft-origin comparisons. Among edited texts (HA versus AH), it reaches 76.68\% AUROC versus 61.26\% for NTS, the strongest baseline. CPD also leads in TPR@5\% on three comparisons. NTS narrowly leads on the fourth, HA versus A. For the same 1,999 humanized AI drafts, replacing unedited human controls with AI-edited human controls requires a higher threshold to stay within 5\% FPR. Recall falls from 73.74\% to 45.32\% (Appendix~\ref{app:hart_source_results}).

\paragraph{Variation across domains and editors.}
We restrict the domain and editor comparisons to texts of 300--500 Falcon-7B tokens before truncation. For draft-origin detection (H/HA versus A/AH), CPD leads on both metrics in all four domains, with AUROC ranging from 85.39\% on essays to 95.45\% on arXiv (Appendix~\ref{app:hart_source_domain_results}). With unedited human controls (H versus AH), rankings vary by editor. CPD leads on both metrics for model and tool rewrites. On drafts rewritten by people, CPD and Binoculars have similar AUROCs, while Fast-DetectGPT achieves higher recall at 5\% FPR than CPD (80.20\% versus 69.31\%, Appendix~\ref{app:hart_humanization_types}).

\subsection{Detection under AI-mediated Rewriting}
\label{sec:gen_detectrlx_ai_rewriting}

\begin{figure*}[t]
    \centering
    \includegraphics[width=0.98\textwidth]{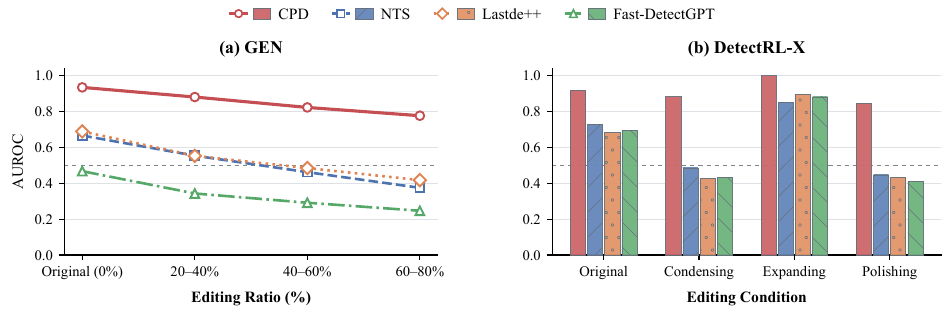}
    \caption{Draft-origin detection under AI-mediated rewriting (AUROC, 0--1). (a) On GEN, the same 936 human drafts serve as negatives before and after AI editing, against 936 fixed AI-generated positives. Original (0\%) uses the unedited drafts. (b) On DetectRL-X, three editing operations modify the same AI drafts, which remain positive against the same human controls. Score directions remain fixed.}
\label{fig:gen_detectrlx_edit}
\end{figure*}

The HART experiment regroups the same scores. Here we keep origin labels fixed and follow drafts across revisions. GEN provides originals and AI-edited versions of 936 human drafts distinct from the direct-detection set. Each draft has one revision per edit-ratio range, with equal representation of the three editors in each range and the same AI controls throughout. On DetectRL-X, condensation, expansion, and polishing change the wording and length of the same 6,000 AI drafts, which we score against fixed human controls. Both experiments preserve domain composition, detector settings, and score directions (Appendix~\ref{app:ai_rewriting_results}).

\paragraph{AI editing of human-origin text.}
On GEN, both metrics decline as the edit ratio increases, while CPD leads at every editing level (Figure~\ref{fig:gen_detectrlx_edit}(a) and Table~\ref{tab:gen_edit_robustness}). In the 60--80\% edit-ratio range, CPD reaches 77.71\% AUROC and all seven baselines fall below 50\% in their fixed score directions. Recall at 5\% FPR falls from 79.06\% on the original drafts to 52.35\% at this highest editing level. Because the AI references are fixed, this decline shows the cost of controlling false positives on increasingly edited human drafts.

\paragraph{AI refinement of AI-origin text.}
On DetectRL-X, CPD leads on both metrics under each rewriting operation (Figure~\ref{fig:gen_detectrlx_edit}(b) and Table~\ref{tab:aa_robustness}). Recall at 5\% FPR is 65.60\% on the original drafts and 65.57\% for rewrites pooled across all three operations. Against the same unedited human controls, CPD's recall rises to 99.90\% after expansion and falls to 53.63\% after condensation and 44.93\% after polishing.

\subsection{Component Analysis}
\label{sec:ablation}

\paragraph{Expectation centering and normalization.}
To isolate what centering contributes, we first average the gains over the same $m$ token positions in each document. At $k=1$, subtracting the expected gain before averaging raises AUROC by 14.6--26.9 percentage points across the three benchmarks ($\bar{\Delta}_k$ versus $\bar{\delta}_k$ in Tables~\ref{tab:component_alternative_statistics} and~\ref{tab:component_decomposition}). Centering also raises AUROC by 32.78--35.81 points when both sums use CPD's variance scale. Once the gains are centered, replacing that scale with $\sqrt m$ (Sqrt) changes AUROC by less than 0.1 point on every benchmark (Table~\ref{tab:component_ablation}). Of these two design choices, subtracting the prefix-specific expectation makes the larger contribution to document ranking.

The analytical scale accounts for variation in contextual gains under the scorer even among texts of equal length (Eq.~\eqref{eq:cpd_variance}). To test how scaling affects threshold transfer, we set each threshold on GEN development data and hold it fixed across the three tests. The resulting FPR ranges across benchmarks are 8.02 percentage points for the mean centered score, 3.28 points for Sqrt, and 3.45 points for CPD. CPD and Sqrt thus yield comparable FPR stability, with different recall trade-offs across benchmarks (Table~\ref{tab:normalization_threshold_transfer}).

\begin{table}[t]
\caption{
Component ablation at $k=1$ (AUROC / TPR@5\%, in \%).
Exp. and Var. denote expectation centering and variance normalization,
respectively. The CPD row includes both components. Bold marks the best result for each metric within each benchmark.
}
\label{tab:component_ablation}
\centering
\small
\setlength{\tabcolsep}{2pt}
\renewcommand{\arraystretch}{0.90}
\begin{tabular*}{\columnwidth}{@{\extracolsep{\fill}}lccc@{}}
\toprule
Method & GEN & DetectRL-X & HART \\
\midrule
CPD
& \textbf{94.18} / \textbf{80.25}
& \textbf{91.90} / \textbf{65.60}
& \textbf{98.31} / \textbf{92.95} \\
w/o Exp.
& 58.37 / 16.25
& 58.81 / 7.58
& 65.53 / 7.45 \\
w/o Var.
& 94.16 / \textbf{80.25}
& 91.85 / 64.55
& 98.22 / 92.85 \\
w/o Both
& 57.97 / 16.25
& 57.86 / 8.83
& 64.67 / 10.05 \\
\bottomrule
\end{tabular*}
\end{table}

\begin{table}[t]
\caption{Component decomposition for H vs.\ A (AUROC in \%). Scores average matched positions without variance normalization. References are the preceding token (rest), a uniformly sampled past token (shuff), or the GEN development marginal. $g=\bar S_{\mathrm A}-\bar S_{\mathrm H}$ is the class mean gap, $s_{\mathrm{pool}}$ the pooled within-class standard deviation, and $J=|g|/s_{\mathrm{pool}}$ expresses the absolute difference between the two class means in units of the pooled standard deviation.}
\label{tab:component_decomposition}
\centering
\small
\setlength{\tabcolsep}{2pt}
\renewcommand{\arraystretch}{0.9}
\begin{tabular*}{\columnwidth}{@{\extracolsep{\fill}}llcccc@{}}
\toprule
Dataset & Component & AUROC & $g$ & $s_{\mathrm{pool}}$ & $J$ \\
\midrule
\multirow{7}{*}{GEN}
& $\bar{\ell}^{\mathrm{full}}$
& 42.10 & $-0.056$ & 0.207 & 0.270 \\
& $-\bar{\ell}^{\mathrm{rest}}$
& 92.84 & $+0.444$ & 0.212 & 2.088 \\
& $\bar{\delta}_k$
& \textbf{95.24} & $+0.388$ & \textbf{0.168} & \textbf{2.312} \\
& $-\bar{\ell}^{\mathrm{marg}}$
& 92.11 & $+0.434$ & 0.198 & 2.188 \\
& $\bar{\delta}_{\mathrm{marg}}$
& 93.04 & $+0.378$ & 0.187 & 2.020 \\
& $-\bar{\ell}^{\mathrm{shuff}}$
& 92.91 & $+0.460$ & 0.210 & 2.186 \\
& $\bar{\delta}_{\mathrm{shuff}}$
& 93.98 & $+0.403$ & 0.187 & 2.156 \\
\midrule
\multirow{7}{*}{DetectRL-X}
& $\bar{\ell}^{\mathrm{full}}$
& 72.21 & $+0.129$ & 0.162 & 0.796 \\
& $-\bar{\ell}^{\mathrm{rest}}$
& 71.87 & $+0.132$ & 0.167 & 0.791 \\
& $\bar{\delta}_k$
& \textbf{91.76} & $+0.260$ & \textbf{0.135} & \textbf{1.935} \\
& $-\bar{\ell}^{\mathrm{marg}}$
& 71.84 & $+0.121$ & 0.151 & 0.799 \\
& $\bar{\delta}_{\mathrm{marg}}$
& 90.03 & $+0.249$ & 0.139 & 1.799 \\
& $-\bar{\ell}^{\mathrm{shuff}}$
& 75.26 & $+0.150$ & 0.157 & 0.957 \\
& $\bar{\delta}_{\mathrm{shuff}}$
& 90.72 & $+0.279$ & 0.149 & 1.866 \\
\midrule
\multirow{7}{*}{HART}
& $\bar{\ell}^{\mathrm{full}}$
& 87.93 & $+0.315$ & 0.219 & 1.439 \\
& $-\bar{\ell}^{\mathrm{rest}}$
& 65.35 & $+0.158$ & 0.241 & 0.653 \\
& $\bar{\delta}_k$
& \textbf{98.26} & $+0.472$ & \textbf{0.147} & \textbf{3.203} \\
& $-\bar{\ell}^{\mathrm{marg}}$
& 69.56 & $+0.164$ & 0.212 & 0.777 \\
& $\bar{\delta}_{\mathrm{marg}}$
& 97.53 & $+0.479$ & 0.155 & 3.089 \\
& $-\bar{\ell}^{\mathrm{shuff}}$
& 71.43 & $+0.184$ & 0.223 & 0.823 \\
& $\bar{\delta}_{\mathrm{shuff}}$
& 97.51 & $+0.499$ & 0.164 & 3.035 \\
\bottomrule
\end{tabular*}
\end{table}

\paragraph{Decomposing the centered signal.}
The centered gain in Eq.~\eqref{eq:response_deviation} is the difference between the full- and restricted-context log-likelihood residuals for the observed token $x_t$
\begin{equation}
\label{eq:residual_components}
\begin{aligned}
\ell_t^{\mathrm{full}}(x_t)
&=
\log p_t(x_t)
-
\mathbb{E}_{v\sim p_t}[\log p_t(v)],\\
\ell_{t,k}^{\mathrm{rest}}(x_t)
&=
\log q_{t,k}(x_t)
-
\mathbb{E}_{v\sim p_t}[\log q_{t,k}(v)].
\end{aligned}
\end{equation}

Both residuals are centered under the full-context distribution $p_t$. The full-context residual also appears in the numerator of Single-Fast~\citep{fastdetectgpt}. To isolate the short prefix's contribution, we average the signed terms over the same $m$ positions without variance normalization. On DetectRL-X, $\bar{\ell}^{\mathrm{full}}$ and $-\bar{\ell}^{\mathrm{rest}}$ give 72.21\% and 71.87\% AUROC, respectively. Their sum, $\bar{\delta}_k$, reaches 91.76\% (Table~\ref{tab:component_decomposition}). The paired score also achieves higher AUROC than either residual alone on the GEN and HART benchmarks.

On all three benchmarks, subtracting the restricted-context residual from the full-context residual cancels some shared within-class variation. The difference has a smaller pooled within-class standard deviation than either residual alone. On GEN, adding the full-context term even narrows the AI--human mean gap from 0.444 to 0.388. The drop in pooled standard deviation from 0.212 to 0.168 is large enough to increase standardized separation $J$, and AUROC rises from 92.84\% to 95.24\%. On DetectRL-X and HART, pairing both widens the mean gap and reduces within-class variation.

\paragraph{Choice of reference context.}
Even a single reference distribution shared by all token positions helps when paired with the full prefix. We obtain this reference, $q_0$, by averaging Falcon-7B's full-context predictions over token positions in the GEN development set. The resulting paired score has higher AUROC than either residual alone on all three benchmarks. Using the immediately preceding token as the short prefix improves AUROC further (Table~\ref{tab:component_decomposition}).

To test adjacency, we compare the preceding token with one drawn uniformly from earlier positions in the same text. Used alone, the random-prefix residual $-\bar{\ell}^{\mathrm{shuff}}$ has higher AUROC on all three benchmarks. Pairing each with the same full-context residual reverses the ordering. Despite a smaller class mean gap, the adjacent-token contrast has lower within-class variation and leads in AUROC by 0.75--1.26 points. We also vary the window length while scoring the same positions ($t=9,\ldots,T$). As $k$ increases from 1 to 2, 4, and 8, CPD's GEN development AUROC falls from 93.84\% to 87.72\%, with the same ordering on all three test sets. CPD outperforms the mean raw gain $\bar{\Delta}_k$ at every window, and both scores peak at $k=1$ (Table~\ref{tab:context_sensitivity}).

\paragraph{Component effects under editing.}
The HART component ablation connects centering to the draft-origin results in Section~\ref{sec:Source_hart}. When both groups are edited (HA versus AH, 2,000 texts per group), the raw gain at CPD's variance scale gives 51.18\% AUROC. Subtracting the expected gain at the same scale raises AUROC to 76.66\% and TPR@5\% from 3.10\% to 45.30\%. Centering improves AUROC on all three editing tasks, while variance normalization has smaller, task-dependent effects (Table~\ref{tab:hart_component_ablation}). Which residual is stronger alone depends on which drafts were edited. The full-context residual leads on HA versus A, and the signed restricted-context residual leads on H versus AH. Their pairing improves AUROC over either alone and reduces within-class variation on all three tasks (Table~\ref{tab:hart_component_decomposition}).

\subsection{Robustness and Generalization}
\label{sec:robustness_generalization}

Section~\ref{sec:ablation} isolates the benefit of pairing contexts. We now test how that contrast separates human and AI texts as the passage and the scoring model change.

\paragraph{Input-length constraints.}
With the reference window fixed at $k=1$, extending prefixes of the same documents from 64 to 256 tokens raises CPD's macro TPR@5\% from 39.90\% to 80.05\%. A longer input contributes more token deviations to the document score, and each added token has a longer full prefix. CPD leads both macro metrics at all three lengths, although Fast-DetectGPT leads in recall on HART at 64 and 128 tokens (Tables~\ref{tab:length_summary} and~\ref{tab:length_control}).

\paragraph{Sensitivity to Character-level Evasion.}
On English DetectRL-X, CPD leads in AUROC under all four character edits (Figure~\ref{fig:char_robustness}). Against unchanged human controls, recall at 5\% FPR falls from 65.60\% on the original AI texts to 31.00\% after ordinary character insertion and 21.38\% after deletion. Zero-width insertion raises recall to 73.50\%.

\paragraph{Cross-lingual generalization.}
With Falcon-7B, CPD leads in AUROC in six of eight languages. TempTest leads in Portuguese and Russian, and CPD's macro lead over TempTest shrinks from 2.82 to 0.49 points without English. At the threshold selected on English GEN, FPR exceeds 5\% in seven languages. Spanish has the highest FPR (11.82\%), although CPD achieves the best AUROC among all detectors on the same Spanish texts (88.80\%, Table~\ref{tab:cross_lingual}).

\paragraph{Cross-backbone generalization.}
On English GEN, CPD leads all five single-model baselines in AUROC and TPR@5\% with each of the five scoring backbones (Table~\ref{tab:cross_backbone_detailed}). Switching from Falcon-7B to Falcon-7B-Instruct lowers CPD's AUROC from 94.18\% to 80.41\%. On the same texts, the mean score for human drafts rises from 0.06 to 1.49, while the mean for AI drafts changes only from 4.57 to 4.62 (Appendix~\ref{app:cross_backbone_results}).

\paragraph{Computational cost.}
Scoring two contexts with one model places CPD between simple single-model statistics and dual-model detectors in both time and memory. On 200 GEN texts (one RTX A6000), CPD used 0.176\,s/text and 15.80\,GB peak memory. Fast-DetectGPT used 0.233\,s/text and 28.19\,GB, and the simpler baselines used 0.113--0.122\,s/text and 14.21\,GB (Appendix~\ref{app:efficiency_results}).

\section{Conclusion}
\label{sec:conclusion}

CPD compares the log-probability gain each observed token receives from earlier context with the gain a single scoring model expects at that prefix. Across three direct-detection benchmarks, subtracting this expectation improves document ranking even under simple averaging. Taking the difference between the centered full- and short-context log likelihoods reduces within-class variation and yields higher AUROC than either term alone. Centering also helps distinguish draft origins after editing. In the HART ablation with both groups edited, it raises draft-origin AUROC from 51.18\% to 76.66\% at a fixed variance scale (Table~\ref{tab:hart_component_ablation}). Across original and edited HART passages, CPD scores track draft origin more closely than the benchmark's final-expression labels. On GEN, heavier AI editing of the same human drafts weakens their separation from direct AI output. On DetectRL-X, pooled recall at 5\% FPR barely changes after rewriting, yet recall on the same AI drafts rises to 99.90\% after expansion and falls to 44.93\% after polishing.

\section*{Limitations}

(1) \textbf{Editing robustness.} On GEN, CPD separates human drafts from a fixed set of direct AI outputs with 93.51\% AUROC before editing. After heavy AI editing of the same human drafts (60--80\% edit ratio), AUROC falls to 77.71\% (Table~\ref{tab:gen_edit_robustness}). On English DetectRL-X, deleting characters from AI texts reduces recall at 5\% FPR from 65.60\% to 21.38\% against unchanged human controls (Figure~\ref{fig:char_robustness}).

(2) \textbf{Generalization.} Our length study uses prefixes of documents at least 256 tokens long, leaving performance on naturally short texts untested. We compare scoring backbones on English GEN and use only Falcon-7B in the multilingual tests, so the effect of backbone choice in other languages is unknown. We evaluate a threshold calibrated on pooled GEN data across benchmarks and languages (Tables~\ref{tab:fixed_threshold} and~\ref{tab:cross_lingual}). Whether a single-domain threshold preserves its target false-positive rate in another domain remains open.

(3) \textbf{Computational efficiency.} CPD uses two full-vocabulary predictions per token. On 200 GEN texts (one RTX A6000), it used 0.176\,s/text and 15.80\,GB peak memory, versus 0.113--0.122\,s/text and 14.21\,GB for simpler single-model statistics (Table~\ref{tab:method_cost}).

\section*{Acknowledgements}

This work was supported in part by the National Natural Science Foundation of China under Grant U2636211.

\section*{AI Use Statement}

We used AI tools to revise research code, identify additional publications for the literature review, polish the writing, and prepare the ICML version of the manuscript.

\section*{Impact Statement}

CPD lets researchers compare how revisions affect detection of draft origin. A human draft can undergo extensive AI editing and still receive a correct human-origin prediction.

A threshold calibrated on English GEN development data to target a 5\% false-positive rate misclassified 11.82\% of human texts in the Spanish DetectRL-X test set as AI-origin (Table~\ref{tab:cross_lingual}). In a school or workplace, a false accusation of AI use can damage a writer's reputation or lead to disciplinary action. Evaluators should validate the threshold on representative texts from human writers in the intended setting. Judgments about authorship or misconduct require human review and evidence independent of the detector.

\bibliographystyle{icml2027}
\bibliography{references}

\twocolumn
\appendix
\raggedbottom
\renewcommand{\dblfloatpagefraction}{0.85}

\section{Datasets and Experimental Settings}
\label{sec:appendix}
\label{sec:dataset_details}

We describe the benchmarks, evaluation protocols, and detector settings. Appendix~\ref{app:mathematical_details} derives the expected deviation and reference variance. Appendix~\ref{app:component_analysis} analyzes the components of CPD and its sensitivity to context length, and Appendix~\ref{app:additional_results} reports further results on all three datasets.

We use 42 as the base random seed for splitting and sampling. Table~\ref{tab:dataset_statistics} gives file counts before class balancing and score filtering. On all three benchmarks, the main comparison uses H and A. The editing analyses also include AI-edited human drafts and humanized or rewritten AI drafts.

\begin{table*}[t]
\caption{Dataset counts before class balancing and valid-score filtering. H denotes unedited human texts. The final column lists the other categories. Each of the 936 GEN drafts has one edited version per edit range. All GEN editing conditions share 936 AI references.}
\label{tab:dataset_statistics}
\centering
\small
\setlength{\tabcolsep}{9pt}
\begin{tabularx}{\linewidth}{@{\extracolsep{\fill}}llrrX@{}}
\toprule
Dataset / condition & Split & H & Other & Other categories \\
\midrule
GEN direct detection & Dev / Test (each) & 2,000 & 2,000 & A \\
GEN editing & Test & 936 & 3,744 & 936 A + 2,808 HA (three edit ranges) \\
DetectRL-X general & English dev & 6,000 & 6,000 & A \\
DetectRL-X general & Test (each language) & 6,000 & 6,000 & A \\
DetectRL-X rewriting & Test (each operation) & 6,000 & 6,000 & AA \\
DetectRL-X rewriting & Test (pooled) & 6,000 & 18,000 & Condensing, expanding, and polishing variants \\
DetectRL-X character noise & Test (each operation) & 6,000 & 6,000 & Perturbed A \\
DetectRL-X character noise & Test (pooled) & 6,000 & 24,000 & Four perturbation variants \\
HART & Dev / Test (each) & 2,000 & 6,000 & 2,000 HA + 2,000 AH + 2,000 A \\
\bottomrule
\end{tabularx}
\end{table*}

\subsection{GEN}
\label{sec:dataset_gen}

GEN~\citep{gen} covers arXiv, Reddit, story generation, wikiHow, and Wikipedia. Its generated texts span eight models from four families: Gemma 3 (12B and 27B)~\citep{gemma3}, Qwen2.5 (7B and 72B)~\citep{qwen25}, GPT-OSS (20B and 120B)~\citep{gptoss}, Llama 3.1 (8B)~\citep{llama3}, and Llama 3.3 (70B)~\citep{llama33}. The editors are Llama 3.3-70B, GPT-OSS-120B, and Qwen2.5-72B.

The development and direct-detection test sets each contain 2,000 H and 2,000 A texts, with 400 examples per category per domain. Their prompt groups are disjoint. Generator counts are balanced at 250 texts per model on development data and range from 249 to 251 on the detection test set. The development set is used for window selection and threshold calibration, without detector training.

The editing comparison uses 936 human drafts, each paired by \texttt{seed\_id} with one AI-edited version in each of three edit-ratio ranges. These drafts span 935 prompt groups, all disjoint from the development and direct-detection prompt groups. Original and edited drafts share the same 936 AI references from the direct-detection test set (Appendix~\ref{app:ai_rewriting_results}).

\subsection{DetectRL-X: General and Multilingual Sets}
\label{sec:dataset_detectrlx}

DetectRL-X~\citep{detectrl} includes texts generated by DeepSeek-V3~\citep{deepseekv3}, Gemini-2.5-Flash~\citep{gemini25}, GPT-4o~\citep{gpt4o}, and Qwen-Max~\citep{qwenmax} across six domains: news, novels, SEO, web text, Wikipedia, and academic writing. We preserve the official train/test split and sample 6,000 paired English examples from the 9,600 available training pairs as a development set, yielding 6,000 H and 6,000 A texts. Each language-specific test set contains 6,000 H and 6,000 A texts and is retained in full. The English test set contains 1,000 pairs per domain and 1,500 AI texts per generator. We use English for the main comparison and evaluate multilingual generalization separately on Arabic, Chinese, French, German, Portuguese, Russian, and Spanish.

\subsection{DetectRL-X: Rewriting and Character Perturbations}
\label{sec:dataset_detectrlx_robustness}

The released English DetectRL-X transformations cover the same four generators as the general set. Each operation modifies AI texts and uses the same 6,000 unchanged human controls. The rewriting pool contains 18,000 variants of 6,000 AI drafts, one condensed, one expanded, and one polished version per draft. We exclude decoder paraphrasing and backtranslation. Domain and generator counts match across the three operations (Appendix~\ref{app:ai_rewriting_results}). Each of the four character operations (insertion, deletion, substitution, and zero-width insertion) is applied separately to 6,000 AI texts, yielding a pool of 24,000 variants. We sample from each pool to balance classes at evaluation time (Appendix~\ref{sec:data_eval_settings}).

\subsection{HART}
\label{sec:dataset_hart}

HART~\citep{hart} covers four English domains: essays, arXiv, creative writing, and news. AI-origin texts are generated by GPT-3.5-Turbo~\citep{gpt35turbo}, GPT-4o~\citep{gpt4o}, Gemini-1.5-Pro~\citep{gemini15}, Claude-3.5-Sonnet~\citep{claude35sonnet}, Qwen2.5-72B-Instruct~\citep{qwen25}, and Llama-3.3-70B-Instruct~\citep{llama33}. The first four models also serve as editors, rephrasing human drafts. The humanized AI drafts include revisions by models, tools, and people. 

We denote HART's four text categories~\citep{hart} by H (type-0, unedited human text), HA (type-1, a human draft refined by AI), AH (type-2, a humanized AI draft), and A (type-3, directly AI-generated text). In each mixed category, the first letter identifies the draft's origin. The H in AH denotes humanized expression, which can result from revision by a model, a tool, or a person. HART's Level-2 task compares types 0/1 with types 2/3 (H/HA vs.\ AH/A).

We preserve the official development and test splits, each containing 8,000 texts with 2,000 examples per category. Each domain contributes 500 examples per category. For fixed-threshold evaluation on HART, we use the thresholds selected on the GEN development split (Appendix~\ref{app:threshold_transfer}). Before score filtering, H vs. A and each two-category editing comparison (H vs. AH, HA vs. A, and HA vs. AH) contain 2,000 texts per class. The pooled draft-origin comparison (H/HA vs. A/AH) and expression diagnostic (H/AH vs. A/HA) each contain 4,000 per class. The joint detector evaluation excluded one AH text (Appendix~\ref{app:hart_source_results}).

\subsection{Preprocessing and Evaluation Settings}
\label{sec:data_eval_settings}

\paragraph{Input preparation.}
We score each nonempty document without its prompt or metadata and retain all released character perturbations, including zero-width characters.

\paragraph{Class balancing.}
For each comparison, we sample $\min(N_-,N_+)$ examples per class without replacement, where $N_-$ and $N_+$ are the available negative and positive counts. Each pooled DetectRL-X evaluation uses all 6,000 human controls and 6,000 AI variants drawn from the pooled operations without quotas by transformation. We compute the pooled metrics once from these selected scores.

\paragraph{Tokenization and scoring.}
Inputs are tokenized without adding special tokens and truncated to at most 1,024 tokens. CPD uses $k=1$ and scores from $t=2$ in the main evaluation. The context-length ablation uses $k\in\{1,2,4,8\}$ and a shared scoring start at $t=9$, as described in Appendix~\ref{app:context_window_results}. Texts that are too short for the requested scoring interval or yield invalid scores are omitted by the evaluator; Table~\ref{tab:dataset_statistics} reports data-file counts before such omissions.

\paragraph{Metrics and aggregation.}
AUROC is the area under the empirical ROC curve of the selected test negatives and positives. TPR@5\% is the largest empirical true-positive rate among ROC operating points whose false-positive rate is at most 0.05. It summarizes the test ROC and does not use a development-set threshold. All detection metrics are reported as percentages. Cross-benchmark macro averages weight the three datasets equally. Absolute gains use unrounded metrics and compare CPD with the strongest baseline for each metric, in percentage points.

\paragraph{Fixed-threshold evaluation.}
Appendix~\ref{app:threshold_transfer} assesses threshold transfer by calibrating one threshold per detector on the GEN development split at a target FPR of 5\% and freezing it across all three test sets. We report observed FPR and TPR and compute balanced accuracy (BA) from integer confusion counts, rounding BA to two decimal places.

\paragraph{Efficiency protocol.}
Cross-backbone comparisons use the full GEN test set. Efficiency is measured separately on the same 200 GEN texts (100 per class) for each method, using BF16 precision on a single NVIDIA RTX A6000 after two warm-up texts. Mean processing time excludes model loading, dataset loading, and warm-up; peak allocated GPU memory includes model weights. CPD reuses predictions for identical one-token restricted contexts within each document. Results are reported in Appendix~\ref{app:efficiency_results}.

\subsection{Detector Sources and Configurations}
\label{app:baseline_settings}

We follow the detector definitions and recommended settings in the original papers, using the official repositories for implementation details. We use the same settings on GEN, DetectRL-X, and HART and indicate where an adopted repository default differs from the corresponding paper.

\paragraph{Shared settings.}
We use \path{tiiuae/falcon-7b} and \path{tiiuae/falcon-7b-instruct} in BF16. Plain-text inputs are tokenized without added special tokens and truncated to 1,024 tokens; dual-model methods require identical token IDs. Unless otherwise specified, targets are $t=2,\ldots,T$, distributions use the full vocabulary, and the random seed is 42. Appendix~\ref{sec:data_eval_settings} gives the preprocessing and evaluation settings used for all three benchmarks.

\paragraph{Loglikelihood.}
The token-probability feature follows GLTR~\citep[Section~2]{gltr}, with mean log-probability aggregation as used for the likelihood baseline in \citet[Section~3.1]{fastdetectgpt}.\footnote{\url{https://github.com/HendrikStrobelt/detecting-fake-text}}
We average the log probability Falcon-7B assigns to each observed token given its full prefix. This score has no additional hyperparameters.

\paragraph{LRR.}
We follow the DetectLLM-LRR definition in \citet[Section~3.1]{detectllm} and its perturbation-free evaluation in that paper's Section~4.\footnote{\url{https://github.com/mbzuai-nlp/DetectLLM}}
Using Falcon-7B, we divide the negative mean token log probability by the mean log rank. Ranks start at one and cover the full vocabulary. LRR requires no perturbation or sampling hyperparameters.

\paragraph{TempTest.}
We adopt the scoring temperature of 0.8 used in the white- and black-box experiments of \citet[Sections~5.2 and~5.4]{temptest}.\footnote{\url{https://github.com/TMKempton/TempTest}}
We use Falcon-7B as the shared scoring backbone, keep this assumed generation temperature fixed across benchmarks, and negate the native statistic to follow the higher-is-AI score convention.

\paragraph{NTS.}
We follow the Falcon-7B configuration and low/high temperatures recommended by \citet[Section~4.1, with hyperparameter analysis in Section~4.5]{nts}, setting them to 0.7 and 1.4, respectively.\footnote{\url{https://github.com/Shixuan-Ma/NTS}}
Sensitivity is the absolute difference between the two document-mean observed-token log probabilities. Following the official implementation, we use the original temperature-1 distribution to compute each token-wise expectation and standard deviation. We center the score by the mean of the expectations and normalize by the mean of the standard deviations.

\paragraph{Fast-DetectGPT.}
We use the conditional-curvature detector and analytical solution of \citet[Section~2.3 and Appendix~B]{fastdetectgpt}, adapting the two-surrogate configuration in that paper's black-box experiments (Section~3.2) to the shared Falcon backbones.\footnote{\url{https://github.com/baoguangsheng/fast-detect-gpt}}
Falcon-7B supplies the sampling/reference distribution, while Falcon-7B-Instruct is the scoring model. Reference expectations and variances are computed analytically over the vocabulary, without Monte Carlo sampling. The single-model comparison uses Falcon-7B for both roles (Appendix~\ref{app:tcd_fast_comparison}).

\paragraph{Binoculars.}
We follow the default Falcon model pair and detection score of \citet[Section~3.3 and Appendix~A.3]{binoculars}.\footnote{\url{https://github.com/ahans30/Binoculars}}
As in the official implementation, Falcon-7B is the observer and Falcon-7B-Instruct the performer. The score is the negated ratio of performer observed-token negative log likelihood to observer--performer cross entropy, evaluated over matched target positions. For threshold transfer, we calibrate on GEN development data as in Appendix~\ref{app:threshold_transfer}.

\paragraph{Lastde++.}
We follow the sampling-normalized Lastde++ detector of \citet[Section~3.2, with experimental settings in Section~4.1]{lastde} and the recommended configuration in its official repository.\footnote{\url{https://github.com/TrustMedia-zju/Lastde_Detector}}
We use Falcon-7B for sampling and Falcon-7B-Instruct for scoring, with 100 conditional token-sequence samples per text at temperature 1. The Std aggregation variant uses embedding size 4, maximum scale 10, and $8m$ histogram bins, where $m$ is the number of scored tokens. The sample count, embedding size, and bin factor follow the paper. We use the official README default of 10 for maximum scale. Section~3.2 of the paper reports 15.

\paragraph{CPD.}
Our detector uses Falcon-7B for both full- and restricted-context predictions, with $k=1$ selected on GEN development data and scoring from $t=2$. Restricted contexts are processed in batches of up to 1,024 windows, reusing identical one-token contexts. Appendices~\ref{app:context_window_results} and~\ref{sec:additional_ablation} detail window selection and component definitions.

\paragraph{TCD-weighted variants.}
We follow the attention-based weighting and dynamic layer selection of \citet{tcd}, using $\tau_{\mathrm{TCD}}=0.7$ as specified in that paper's Section~4.1.\footnote{\url{https://github.com/RecursionM/TCD}}
We adapt the surrogate models to the shared Falcon backbones: Fast+TCD obtains attention weights from Falcon-7B-Instruct, while CPD+TCD uses Falcon-7B. The separate comparison uses eager attention and restricted-context batches of up to 256 windows (Appendix~\ref{app:tcd_fast_comparison}).

\paragraph{Alternative scoring backbones.}
The GEN comparisons additionally use \path{tiiuae/falcon-7b-instruct}, \path{EleutherAI/gpt-neo-2.7B}, \path{meta-llama/Llama-2-7b-hf}, and \path{meta-llama/Llama-3.1-8B}. Each single-model method uses the selected backbone and its tokenizer, with temperature parameters, $k=1$, input length, and scoring interval unchanged.

\paragraph{Consistent score orientation.}
Each baseline's score direction follows its original decision rule. When the native statistic assigns lower values to AI-generated text, we negate it once so that higher scores indicate stronger evidence of AI origin. We fix this direction before testing and retain it across benchmarks and generator subsets.

\paragraph{Interpreting AUROC below 50\%.}
With this convention, AUROC below 50\% means that a randomly selected human text more often outscores a randomly selected AI text than the reverse. On GEN, Falcon-7B log likelihood gives 2.84\% AUROC for GPT-OSS-20B outputs and 78.34\% for GPT-OSS-120B outputs. The ordering thus reverses across these two generator subsets. CPD ranks AI drafts above human texts more often on both subsets, with AUROCs of 78.42\% and 96.44\%, respectively (Table~\ref{tab:gen_generator_results}).

\FloatBarrier

\section{Analytical Derivations}
\label{app:mathematical_details}

\subsection{Interpretation of Expectation Centering}
\label{app:expectation_centering}

At a fixed prefix, the centering term in Eq.~\eqref{eq:response_deviation}
is constant across candidate tokens. Averaging over the source
distribution $r_t$ gives
$\mathbb E_{r_t}[\delta_{t,k}]
=\mathbb E_{r_t}[\Delta_{t,k}]-\mathbb E_{p_t}[\Delta_{t,k}]$.
Writing $\log(p_t(v)/q_{t,k}(v))$ as
$\log(r_t(v)/q_{t,k}(v))-\log(r_t(v)/p_t(v))$ for tokens with
positive probability under $r_t$ gives the following identity
\begin{equation}
\label{eq:appendix_kl_interpretation}
\begin{aligned}
\mathbb E_{v\sim r_t}[\delta_{t,k}(v)]
&=
D_{\mathrm{KL}}(r_t\Vert q_{t,k})\\
&\quad-
D_{\mathrm{KL}}(r_t\Vert p_t)\\
&\quad-
D_{\mathrm{KL}}(p_t\Vert q_{t,k}).
\end{aligned}
\end{equation}

The mean deviation vanishes when $r_t=p_t$ or when the shifts from
$p_t$ to a different $r_t$ cancel in the weighted sum in
Eq.~\eqref{eq:source_expectation}.
For two sources at the same prefix, the common centering term cancels
from the difference between their expected deviations. The ablation in
Section~\ref{sec:ablation} tests how prefix-specific
centering affects document rankings.

\subsection{Analytical Reference Variance}
\label{app:reference_variance}

Condition on the observed text $x$, fixing the full- and restricted-prefix
distributions $p_t$ and $q_{t,k}$ at each scored position. The hypothetical
reference tokens $\tilde{x}_t$ are independent across positions, with
$\tilde{x}_t\sim p_t$. The contextual response
$Y_{t,k}=\Delta_{t,k}(\tilde{x}_t)$ has expectation
\begin{equation}
\label{eq:appendix_reference_mean}
\begin{aligned}
\mathbb E[Y_{t,k}]
&=
\sum_{v\in\mathcal V}
p_t(v)\Delta_{t,k}(v)\\
&=
\mu_{t,k}
=
D_{\mathrm{KL}}(p_t\Vert q_{t,k}).
\end{aligned}
\end{equation}

Subtracting the squared mean from the second moment gives the
token-level reference variance
\begin{equation}
\label{eq:appendix_token_variance}
\begin{aligned}
\operatorname{Var}[Y_{t,k}]
&=
\mathbb E[Y_{t,k}^{2}]
-
\bigl(\mathbb E[Y_{t,k}]\bigr)^2\\
&=
\sum_{v\in\mathcal V}
p_t(v)\Delta_{t,k}(v)^2
-\mu_{t,k}^{2}.
\end{aligned}
\end{equation}

The document-level reference response is
$\Delta_k(\tilde{x};x)
=\sum_{t\in\mathcal I_k}Y_{t,k}$.
Independence across positions makes all covariances between distinct reference responses zero. Summing the individual response variances in Eq.~\eqref{eq:cpd_variance} gives
\begin{equation}
\label{eq:appendix_document_variance}
\begin{aligned}
&\operatorname{Var}[\Delta_k(\tilde{x};x)]\\
&=
\sum_{t\in\mathcal I_k}
\operatorname{Var}[Y_{t,k}]\\
&=
\sum_{t\in\mathcal I_k}
\left(
\sum_{v\in\mathcal V}
p_t(v)\Delta_{t,k}(v)^2
-\mu_{t,k}^{2}
\right).
\end{aligned}
\end{equation}

Taking the square root gives the reference standard deviation
\begin{equation}
\label{eq:appendix_reference_std}
\begin{aligned}
&\sigma_k(x)\\
&=
\Bigg[
\sum_{t\in\mathcal I_k}
\Bigg(
\sum_{v\in\mathcal V}
p_t(v)\Delta_{t,k}(v)^2
-\mu_{t,k}^{2}
\Bigg)
\Bigg]^{1/2}.
\end{aligned}
\end{equation}

The vocabulary sums give the reference moments directly from the
scorer's predictions, without sampling alternative tokens or passages.
CPD uses the numerically stabilized denominator
$\sqrt{\sigma_k(x)^2+\epsilon}$ in Eq.~\eqref{eq:cpd}.

\section{Component and Sensitivity Analyses}
\label{app:component_analysis}

The ablations in Section~\ref{sec:ablation} vary expectation centering and the document-level scale. We use the same aggregation rule to compare the full- and restricted-context residuals.

\subsection{Ablation Definitions and Alternative Statistics}
\label{sec:additional_ablation}

\paragraph{Shared scoring protocol.}
CPD, its three ablations, and the three alternative statistics use the same scoring model and $k=1$, with common target positions $t=2,\ldots,T$ after truncation to at most 1,024 tokens. We fix score directions before testing, with higher scores indicating stronger evidence of AI origin. Each test ROC curve gives AUROC and TPR@5\%, the highest TPR at FPR $\leq 5\%$.

\paragraph{Component definitions.}
Let $V_k=\operatorname{Var}[\Delta_k(\tilde{x};x)]$ and $m=|\mathcal I_k|$. For contextual response $\Delta_k(x)$ with expectation $\mu_k(x)$ (Section~\ref{sec:method}), the direct-detection and editing experiments in Tables~\ref{tab:component_ablation} and~\ref{tab:hart_component_ablation} compare the following scores
\begin{equation}
\label{eq:component_ablation_scores}
\begin{aligned}
\mathrm{CPD} &= \frac{\Delta_k(x)-\mu_k(x)}{\sqrt{V_k+\epsilon}},\\
\text{w/o Exp.} &= \Delta_k(x)/\sqrt{V_k+\epsilon},\\
\text{w/o Var.} &= \frac{\Delta_k(x)-\mu_k(x)}{\sqrt{m}},\\
\text{w/o Both} &= \Delta_k(x)/\sqrt{m}.
\end{aligned}
\end{equation}
Both variants without variance normalization divide the sum by the square root of the number of scored tokens.

\paragraph{Alternative statistics.}
We also evaluate the average contextual response, its average model-expected value, and average restricted-context negative log-likelihood (Res-NLL). Overbars denote means over matched target positions.
\begin{equation}
\label{eq:alternative_statistics}
\begin{aligned}
\bar\Delta_k(x) &= \Delta_k(x)/m,\\
\bar\mu_k(x) &= \mu_k(x)/m\\
&= \frac{1}{m}\sum_{t\in\mathcal I_k}
D_{\mathrm{KL}}(p_t\Vert q_{t,k}),\\
\mathrm{Res\text{-}NLL}_k(x) &= -\frac{1}{m}\sum_{t\in\mathcal I_k}
\log q_{t,k}(x_t).
\end{aligned}
\end{equation}
On the H versus A test sets, CPD exceeds all three alternative statistics in both AUROC and TPR@5\% (Tables~\ref{tab:component_ablation} and~\ref{tab:component_alternative_statistics}). To isolate centering, we compare the mean raw gain $\bar\Delta_1$ with the mean centered gain $\bar\delta_1=\bar\Delta_1-\bar\mu_1$ in Table~\ref{tab:component_decomposition}. Both average the same positions with weight $1/m$. Subtracting the expected gain raises AUROC by 14.6--26.9 percentage points across the three benchmarks.

\begin{table}[!htbp]
\caption{Alternative statistics at $k=1$ for H versus A (AUROC / TPR@5\%, in \%). Overbars denote means over matched target positions. $\bar\mu_k$ averages the KL divergence from the full-context to the restricted-context distribution. $\mathrm{Res\text{-}NLL}_k$ is mean restricted-context negative log-likelihood. Bold marks the best unrounded result among the three statistics for each metric and benchmark.}
\label{tab:component_alternative_statistics}
\centering
\small
\begin{tabular*}{\columnwidth}{@{\extracolsep{\fill}}lccc@{}}
\toprule
Statistic & GEN & DetectRL-X & HART \\
\midrule
$\bar\Delta_k$ & 68.36 / 13.45 & \textbf{71.04} / 3.38 & \textbf{83.62} / 19.45 \\
$\bar\mu_k$ & 49.36 / 4.90 & 53.31 / 0.85 & 63.92 / 1.35 \\
$\mathrm{Res\text{-}NLL}_k$ & \textbf{76.89} / \textbf{29.90} & 64.66 / \textbf{4.37} & 67.97 / \textbf{24.35} \\
\bottomrule
\end{tabular*}
\end{table}

\paragraph{Fixed marginal reference.}
The marginal control gives every position the same reference distribution $q_0$, the average of Falcon-7B's full-context predictive distributions over token positions on GEN development data. Substituting $q_0$ for $q_{t,1}$ in Eq.~\eqref{eq:residual_components} defines the marginal residual $\ell_t^{\mathrm{marg}}$. Its centering expectation still uses the position's original full-context distribution $p_t$. Comparing $-\bar\ell^{\mathrm{marg}}$ with the paired score $\bar\delta_{\mathrm{marg}}=\bar\ell^{\mathrm{full}}-\bar\ell^{\mathrm{marg}}$ tests how much pairing helps when the reference has no access to the preceding token.

\paragraph{Random one-token reference.}
For each target $x_t$, $t=2,\ldots,T$, we draw an index $j_t$ uniformly from $\{1,\ldots,t-1\}$ and use $q_t^{\mathrm{shuff}}(v)=P_\theta(v\mid x_{j_t})$. The immediate predecessor is eligible and is the only choice at $t=2$. We reset the random seed to 42 for each document and use the same draws for both rows. Replacing $q_{t,1}$ with $q_t^{\mathrm{shuff}}$ in Eq.~\eqref{eq:residual_components} defines $\ell_t^{\mathrm{shuff}}$, with its expectation still taken under the original $p_t$. Table~\ref{tab:component_decomposition} reports its signed document mean $-\bar{\ell}^{\mathrm{shuff}}$ and the mean centered contrast $\bar{\delta}_{\mathrm{shuff}}=\bar{\ell}^{\mathrm{full}}-\bar{\ell}^{\mathrm{shuff}}$.

\paragraph{Class-conditional separation analysis.}
Tables~\ref{tab:component_decomposition} and~\ref{tab:hart_component_decomposition} compare the signed terms $\bar\ell^{\mathrm{full}}$ and $-\bar\ell^{\mathrm{rest}}$ with their sum $\bar\delta_k$ (Eq.~\eqref{eq:residual_components}). Each mean assigns weight $1/m$ to the same $m$ scored positions. Equation~\eqref{eq:component_ablation_scores} gives the corresponding variance and square-root length scales for the summed gains. For each document-level score $S$, let $-$ and $+$ denote the negative and positive classes. We report the signed class mean gap $g_S=\bar S_+-\bar S_-$ and standardized separation $J(S)=|g_S|/s_{\mathrm{pool}}(S)$, where
\begin{equation}
\label{eq:pooled_class_sd}
s_{\mathrm{pool}}(S)=
\sqrt{\frac{(n_--1)s_-^2+(n_+-1)s_+^2}{n_-+n_+-2}},
\end{equation}
with $n_c$ and $s_c^2$ denoting the sample count and unbiased sample variance for class $c$. In direct detection, H is negative and A positive, with 2,000 documents per class on GEN and HART and 6,000 per class on DetectRL-X. Appendix~\ref{app:hart_component_results} specifies the class pairs for edited texts.

On GEN, the marginal control illustrates why AUROC and $J$ can move in different directions. Pairing the marginal residual with the full-context residual raises AUROC from 92.11\% to 93.04\%, while $J$ falls from 2.188 to 2.020 (Table~\ref{tab:component_decomposition}). AUROC depends on the ordering of human and AI document scores. $J$ depends only on the class means and pooled within-class spread. The adjacent and random one-token references show a different pattern. Switching from the random to the adjacent token in the paired score narrows the class mean gap from 0.403 to 0.388, while reducing $s_{\mathrm{pool}}$ from 0.187 to 0.168. The reduction in spread more than offsets the smaller gap in the standardized measure.

\subsection{HART Component Ablations under Editing}
\label{app:hart_component_results}

We evaluate HA versus A, H versus AH, and HA versus AH on the HART test set. Within each task, we recompute the four scores in Eq.~\eqref{eq:component_ablation_scores} at $k=1$ on the same 2,000 texts per class. Each comparison lists the human-origin group first (negative) and the AI-origin group second (positive). In both comparisons involving AH, the component runs retain all 2,000 AH texts. The joint evaluation across detectors retained 1,999 after score filtering.

\begin{table}[!htbp]
\caption{HART component ablations under editing at $k=1$ (AUROC / TPR@5\%, in \%). Each task has 2,000 texts per class, with the first category negative and the second positive. Exp. and Var. denote expectation centering and variance normalization. Best values for each metric and task, selected before rounding, are bold.}
\label{tab:hart_component_ablation}
\centering
\small
\begin{tabular*}{\columnwidth}{@{\extracolsep{\fill}}lccc@{}}
\toprule
Method & HA vs. A & H vs. AH & HA vs. AH \\
\midrule
CPD
& \textbf{90.82} / \textbf{65.90}
& 90.79 / 73.70
& 76.66 / \textbf{45.30} \\
w/o Exp.
& 60.50 / 7.65
& 56.92 / 2.90
& 51.18 / 3.10 \\
w/o Var.
& 90.08 / 62.90
& \textbf{90.90} / \textbf{74.50}
& \textbf{76.87} / 44.60 \\
w/o Both
& 59.32 / 7.20
& 58.73 / 6.45
& 53.35 / 4.60 \\
\bottomrule
\end{tabular*}
\end{table}

At CPD's variance scale, subtracting the expected gain raises AUROC by 25.48--33.87 percentage points across the three editing tasks (Table~\ref{tab:hart_component_ablation}). With centered gains, the variance scale gives higher AUROC and TPR@5\% on HA versus A, while $\sqrt{m}$ gives higher values on H versus AH. When both classes are edited, using the variance scale in place of $\sqrt{m}$ raises TPR@5\% from 44.60\% to 45.30\% and lowers AUROC from 76.87\% to 76.66\%.

\subsection{HART Component Decomposition under Editing}
\label{app:hart_component_decomposition}

We apply the residual decomposition from Table~\ref{tab:component_decomposition} to the same documents and token positions as the HART editing ablations. Table~\ref{tab:hart_component_decomposition} compares the signed document means $\bar\ell^{\mathrm{full}}$ and $-\bar\ell^{\mathrm{rest}}$ with their sum $\bar\delta_k$, using $1/m$ aggregation throughout. We measure within-class spread and standardized mean separation as in Eq.~\eqref{eq:pooled_class_sd}.

\begin{table}[!htbp]
\caption{HART component decomposition under editing at $k=1$ (AUROC in \%). Scores average matched target positions without variance normalization. Each task has 2,000 texts per class, with negatives listed first. The mean gap $g$ is the positive class mean minus the negative class mean, and $J=|g|/s_{\mathrm{pool}}$. Within each task, bold marks the highest AUROC and $J$ and the lowest $s_{\mathrm{pool}}$.}
\label{tab:hart_component_decomposition}
\centering
\small
\setlength{\tabcolsep}{4pt}
\begin{tabular*}{\columnwidth}{@{\extracolsep{\fill}}llcccc@{}}
\toprule
Task & Component & AUROC & $g$ & $s_{\mathrm{pool}}$ & $J$ \\
\midrule
\multirow{3}{*}{HA vs. A}
& $\bar{\ell}^{\mathrm{full}}$
& 81.93 & $+0.216$ & 0.224 & 0.965 \\
& $-\bar{\ell}^{\mathrm{rest}}$
& 58.41 & $+0.110$ & 0.256 & 0.429 \\
& $\bar{\delta}_k$
& \textbf{91.05} & $+0.326$ & \textbf{0.169} & \textbf{1.927} \\
\midrule
\multirow{3}{*}{H vs. AH}
& $\bar{\ell}^{\mathrm{full}}$
& 55.41 & $+0.007$ & 0.226 & 0.031 \\
& $-\bar{\ell}^{\mathrm{rest}}$
& 82.76 & $+0.356$ & 0.271 & 1.313 \\
& $\bar{\delta}_k$
& \textbf{90.90} & $+0.363$ & \textbf{0.185} & \textbf{1.966} \\
\midrule
\multirow{3}{*}{HA vs. AH}
& $\bar{\ell}^{\mathrm{full}}$
& 42.68 & $-0.092$ & 0.231 & 0.398 \\
& $-\bar{\ell}^{\mathrm{rest}}$
& 77.07 & $+0.308$ & 0.284 & \textbf{1.084} \\
& $\bar{\delta}_k$
& \textbf{77.66} & $+0.216$ & \textbf{0.202} & 1.069 \\
\bottomrule
\end{tabular*}
\end{table}

The full-context residual has higher AUROC on HA versus A, and the signed restricted-context residual has higher AUROC on H versus AH. Their sum yields the highest AUROC and lowest pooled within-class standard deviation on all three tasks. It also yields the largest $J$ in the two comparisons with one edited group.

On HA versus AH, the full-context component's negative mean gap ($-0.092$) partially offsets the signed restricted-context gap ($+0.308$), giving a combined gap of $+0.216$. The mean gap shrinks proportionally more than the pooled within-class standard deviation, which falls from 0.284 to 0.202. As a result, $J$ falls slightly from 1.084 to 1.069, even as AUROC rises from 77.07\% to 77.66\%.

\subsection{Reference Standard Deviation and Threshold Transfer}
\label{app:normalization_threshold_transfer}

\paragraph{Role of the reference scale.}
Table~\ref{tab:normalization_threshold_transfer} compares three scales for $\delta_k(x)=\Delta_k(x)-\mu_k(x)$. Mean divides by the number of scored positions $m$, and Sqrt divides by $\sqrt{m}$. CPD also accounts for the model's variance at each position through the total reference variance $V_k=\sigma_k(x)^2$ (Eq.~\eqref{eq:appendix_reference_std}). In terms of the mean deviation $\bar\delta_k(x)=\delta_k(x)/m$, CPD is
\begin{equation}
\label{eq:cpd_normalized_mean}
\operatorname{CPD}_k(x)
=\frac{\bar\delta_k(x)}{\sqrt{\sigma_k(x)^2+\epsilon}/m}.
\end{equation}
Under independent reference sampling with fixed contexts, $\sigma_k(x)/m$ is the reference standard deviation of the mean contextual response. The residual decomposition instead estimates $s_{\mathrm{pool}}$ empirically across documents within each class. For numerical stability, the implementation floors $V_k$ at $10^{-8}$ before taking its square root. Using the additive $10^{-8}$ in Eq.~\eqref{eq:cpd} gives the same development and test classifications after calibration.

\begin{table}[!htbp]
\caption{Test FPR / TPR (\%) for H versus A at $k=1$. Each normalization uses a threshold selected on GEN development data at a target FPR of 5\% and held fixed across the three test sets.}
\label{tab:normalization_threshold_transfer}
\centering
\small
\setlength{\tabcolsep}{2pt}
\begin{tabular*}{\linewidth}{@{\extracolsep{\fill}}lccc@{}}
\toprule
Score & GEN & DetectRL-X & HART \\
\midrule
Mean $(\delta_k/m)$ & 3.85 / 83.75 & 11.87 / 79.52 & 5.85 / 94.25 \\
Sqrt $(\delta_k/\sqrt{m})$ & 3.70 / 77.95 & 6.53 / 69.15 & 3.25 / 90.40 \\
CPD $(\delta_k/\sqrt{V_k+\epsilon})$ & 3.75 / 77.70 & 6.95 / 70.35 & 3.50 / 91.65 \\
\bottomrule
\end{tabular*}
\end{table}

\paragraph{Development-set threshold calibration.}
We select a threshold for each score on the same 2,000 human and 2,000 AI texts from GEN development data using Eq.~\eqref{eq:fixed_threshold} (Table~\ref{tab:normalization_dev_thresholds}). Mean and Sqrt use the saved \texttt{R} and \texttt{w/o Variance} columns, respectively. We freeze each threshold at full precision and classify scores at or above it as AI-origin on all three test sets. Each variant scores the same 2,000 texts per class on GEN and HART and 6,000 per class on DetectRL-X (Table~\ref{tab:normalization_threshold_transfer}). Appendix~\ref{app:threshold_transfer} uses the same protocol to compare CPD with the seven baseline detectors.

\begin{table}[!htbp]
\caption{GEN development thresholds for the three normalization choices at $k=1$. FPR and TPR are percentages at a target FPR of 5\%. We round thresholds for display only and test at full precision.}
\label{tab:normalization_dev_thresholds}
\centering
\small
\begin{tabularx}{\columnwidth}{@{\extracolsep{\fill}}lccc@{}}
\toprule
Score & Threshold & Dev FPR & Dev TPR \\
\midrule
Mean & 0.19754306 & 5.00 & 84.60 \\
Sqrt & 4.37745190 & 5.00 & 78.80 \\
CPD  & 2.05796176 & 4.95 & 78.85 \\
\bottomrule
\end{tabularx}
\end{table}

\paragraph{False-positive stability across benchmarks.}
The FPR range in Section~\ref{sec:ablation} is the maximum minus the minimum realized FPR across the three test sets. The largest absolute departure from the development FPR is 2.00 percentage points for CPD, 1.75 for Sqrt, and 6.87 for Mean. Mean has the highest recall and FPR on all three tests. Relative to Sqrt, CPD detects more AI texts on DetectRL-X and HART at slightly higher FPRs. On GEN, Sqrt detects more AI texts and misclassifies fewer human texts than CPD (Table~\ref{tab:normalization_threshold_transfer}).

\subsection{Short-Context Window Size}
\label{app:context_window_results}

\begin{table*}[!tp]
\caption{Window size $k$ (AUROC / TPR@5\%, in \%). We select $k=1$ on GEN development data. All windows score $t=9,\ldots,T$, and Mean $\Delta=\Delta_k(x)/(T-8)$ is the signed mean contextual response. Bold marks the best value over $k$ for each method, split, and metric.}
\label{tab:context_sensitivity}
\centering
\small
\setlength{\tabcolsep}{4pt}
\begin{tabular*}{\linewidth}{@{\extracolsep{\fill}}lcccc@{}}
\toprule
\multirow{2}{*}{Method} & Dev. & \multicolumn{3}{c}{Test.} \\
\cmidrule(lr){2-2}
\cmidrule(lr){3-5}
& GEN & GEN & DetectRL-X & HART \\
\midrule
Mean $\Delta_{k=1}$ & \textbf{67.37} / \textbf{12.90} & \textbf{68.61} / \textbf{13.50} & \textbf{70.28} / \textbf{3.23} & \textbf{83.97} / \textbf{19.20} \\
CPD$_{k=1}$ & \textbf{93.84} / \textbf{79.00} & \textbf{94.34} / \textbf{80.55} & \textbf{91.41} / \textbf{63.78} & \textbf{98.25} / \textbf{93.15} \\
\midrule
Mean $\Delta_{k=2}$ & 60.41 / 8.65 & 62.05 / 9.00 & 60.41 / 1.15 & 78.76 / 8.60 \\
CPD$_{k=2}$ & 91.88 / 74.00 & 92.45 / 76.05 & 90.42 / 60.75 & 97.58 / 90.20 \\
\midrule
Mean $\Delta_{k=4}$ & 55.17 / 5.75 & 57.49 / 6.60 & 53.40 / 1.13 & 72.41 / 11.85 \\
CPD$_{k=4}$ & 89.75 / 68.75 & 90.16 / 69.10 & 88.76 / 59.02 & 96.51 / 86.25 \\
\midrule
Mean $\Delta_{k=8}$ & 57.09 / 4.40 & 58.78 / 6.15 & 52.04 / 1.93 & 72.82 / 15.10 \\
CPD$_{k=8}$ & 87.72 / 63.35 & 88.34 / 65.20 & 86.32 / 52.23 & 95.19 / 81.20 \\
\bottomrule
\end{tabular*}
\end{table*}

\paragraph{Window selection on the development set.}
Among $k\in\{1,2,4,8\}$, the one-token window gives CPD its highest AUROC (93.84\%) and TPR@5\% (79.00\%) on GEN development data. We select $k=1$ on this split and evaluate sensitivity to window length on the three test sets (Table~\ref{tab:context_sensitivity}).

\paragraph{Matched scoring interval.}
All four windows score the same documents and positions $t=9,\ldots,T$, with $m=T-8$ target tokens after truncation to at most 1,024 tokens. The GEN development and test splits and the HART test split each contain 2,000 texts per class, and the DetectRL-X test split contains 6,000 per class. The first eight tokens remain available to the full-context predictions. Mean $\Delta$ is the signed average $\Delta_k(x)/m$ over these positions. The main evaluation at $k=1$ scores from $t=2$.

\paragraph{Effect of context restriction.}
The development-set choice $k=1$ also gives the highest AUROC and TPR@5\% for CPD and Mean $\Delta$ on all three test sets. Across all four splits, CPD leads Mean $\Delta$ on both metrics at every window, and both CPD metrics fall with each increase in $k$. Extending the short prefix from one token to eight lowers CPD's test recall at 5\% FPR by 11.55--15.35 percentage points.

\section{Additional Detection Results}
\label{app:additional_results}

This appendix gives uncertainty estimates, threshold-transfer results, and detailed comparisons for Sections~\ref{sec:main_results}--\ref{sec:robustness_generalization}. Appendix~\ref{app:component_analysis} reports the component ablations.

\subsection{Bootstrap Confidence Intervals}
\label{app:bootstrap_ci}

\begin{table*}[!tp]
\caption{Main detection results for H versus A, with AUROC and TPR@5\% in percent. Below each point estimate is its 95\% percentile interval from 2,000 class-stratified bootstrap replicates. CPD uses $k=1$, and $^\dagger$ marks configurations that use two language models.}
\label{tab:bootstrap_ci}
\centering
\small
\setlength{\tabcolsep}{3pt}
\renewcommand{\arraystretch}{1.0}
\begin{tabular*}{\linewidth}{@{\extracolsep{\fill}}lcccccc@{}}
\toprule
\multirow{2}{*}{Method}
& \multicolumn{2}{c}{GEN}
& \multicolumn{2}{c}{DetectRL-X}
& \multicolumn{2}{c}{HART} \\
\cmidrule(lr){2-3}
\cmidrule(lr){4-5}
\cmidrule(lr){6-7}
& AUROC & TPR@5\%
& AUROC & TPR@5\%
& AUROC & TPR@5\% \\
\midrule
Loglikelihood
& \shortstack{43.04\\{\scriptsize [41.24, 44.87]}}
& \shortstack{8.65\\{\scriptsize [6.75, 11.30]}}
& \shortstack{55.52\\{\scriptsize [54.54, 56.54]}}
& \shortstack{8.72\\{\scriptsize [7.47, 9.80]}}
& \shortstack{75.77\\{\scriptsize [74.34, 77.34]}}
& \shortstack{15.20\\{\scriptsize [8.30, 24.30]}} \\
\addlinespace[2pt]
LRR
& \shortstack{47.94\\{\scriptsize [46.09, 49.77]}}
& \shortstack{13.65\\{\scriptsize [11.85, 15.95]}}
& \shortstack{53.33\\{\scriptsize [52.40, 54.35]}}
& \shortstack{10.57\\{\scriptsize [9.60, 11.58]}}
& \shortstack{81.25\\{\scriptsize [79.91, 82.61]}}
& \shortstack{49.00\\{\scriptsize [44.05, 52.95]}} \\
\addlinespace[2pt]
TempTest
& \shortstack{43.91\\{\scriptsize [42.09, 45.81]}}
& \shortstack{19.90\\{\scriptsize [18.35, 22.10]}}
& \shortstack{72.74\\{\scriptsize [71.88, 73.63]}}
& \shortstack{21.80\\{\scriptsize [20.32, 23.42]}}
& \shortstack{87.69\\{\scriptsize [86.55, 88.95]}}
& \shortstack{74.85\\{\scriptsize [71.95, 77.60]}} \\
\addlinespace[2pt]
NTS
& \shortstack{67.75\\{\scriptsize [66.16, 69.37]}}
& \shortstack{33.55\\{\scriptsize [31.00, 36.35]}}
& \shortstack{72.73\\{\scriptsize [71.85, 73.61]}}
& \shortstack{19.92\\{\scriptsize [18.48, 21.75]}}
& \shortstack{93.30\\{\scriptsize [92.52, 94.13]}}
& \shortstack{82.15\\{\scriptsize [80.25, 84.40]}} \\
\addlinespace[2pt]
Single-Fast
& \shortstack{42.79\\{\scriptsize [40.96, 44.68]}}
& \shortstack{21.30\\{\scriptsize [19.00, 23.10]}}
& \shortstack{72.72\\{\scriptsize [71.85, 73.64]}}
& \shortstack{25.30\\{\scriptsize [23.65, 27.12]}}
& \shortstack{87.71\\{\scriptsize [86.54, 88.99]}}
& \shortstack{76.60\\{\scriptsize [74.55, 78.90]}} \\
\addlinespace[2pt]
Binoculars$^\dagger$
& \shortstack{46.38\\{\scriptsize [44.49, 48.27]}}
& \shortstack{23.80\\{\scriptsize [21.85, 25.90]}}
& \shortstack{68.42\\{\scriptsize [67.48, 69.36]}}
& \shortstack{9.22\\{\scriptsize [8.13, 10.63]}}
& \shortstack{91.09\\{\scriptsize [89.98, 92.20]}}
& \shortstack{85.80\\{\scriptsize [84.40, 87.60]}} \\
\addlinespace[2pt]
Fast-DetectGPT$^\dagger$
& \shortstack{47.04\\{\scriptsize [45.14, 48.91]}}
& \shortstack{26.50\\{\scriptsize [24.20, 28.40]}}
& \shortstack{69.39\\{\scriptsize [68.48, 70.33]}}
& \shortstack{15.70\\{\scriptsize [14.60, 16.98]}}
& \shortstack{90.78\\{\scriptsize [89.69, 91.89]}}
& \shortstack{84.25\\{\scriptsize [82.70, 86.05]}} \\
\addlinespace[2pt]
Lastde++$^\dagger$
& \shortstack{68.52\\{\scriptsize [66.88, 70.28]}}
& \shortstack{43.35\\{\scriptsize [40.45, 45.90]}}
& \shortstack{68.40\\{\scriptsize [67.47, 69.39]}}
& \shortstack{11.85\\{\scriptsize [10.87, 13.25]}}
& \shortstack{93.54\\{\scriptsize [92.64, 94.40]}}
& \shortstack{85.60\\{\scriptsize [84.00, 87.25]}} \\
\midrule
CPD$_{k=1}$ (Ours)
& \shortstack{\textbf{94.18}\\{\scriptsize \textbf{[93.42, 94.86]}}}
& \shortstack{\textbf{80.25}\\{\scriptsize \textbf{[77.70, 82.15]}}}
& \shortstack{\textbf{91.90}\\{\scriptsize \textbf{[91.41, 92.37]}}}
& \shortstack{\textbf{65.60}\\{\scriptsize \textbf{[62.72, 67.53]}}}
& \shortstack{\textbf{98.31}\\{\scriptsize \textbf{[97.96, 98.65]}}}
& \shortstack{\textbf{92.95}\\{\scriptsize \textbf{[91.70, 94.55]}}} \\
\bottomrule
\end{tabular*}
\end{table*}

We treat documents as independent sampling units when estimating uncertainty in Table~\ref{tab:main_results}. Within each benchmark, we resample documents with replacement separately in the human and AI classes, retaining 2,000 documents per class on GEN and HART and 6,000 on DetectRL-X. We draw 2,000 bootstrap replicates with seed 42 and use the same document indices for every detector within a replicate.

For each bootstrap replicate, we record the area under its ROC curve and the highest TPR at an FPR of at most 5\%. The 2.5th and 97.5th percentiles across replicates give the bounds of a 95\% interval for each detector, benchmark, and metric. These intervals are conditional on the saved detector scores and the fixed number of texts in each class (Table~\ref{tab:bootstrap_ci}).

CPD's lower confidence limit exceeds every compared baseline's upper limit for both metrics on each benchmark.

\subsection{Fixed-Threshold Evaluation Across Benchmarks}
\label{app:threshold_transfer}

\begin{table*}[!tp]
\caption{Fixed-threshold detection of unedited human and AI texts (H vs.\ A). For each test set, cells give FPR / TPR / balanced accuracy (BA) in percent. Each detector uses its GEN development threshold (target FPR 5\%) at full precision on all three test sets. We round thresholds only for display. For each benchmark, bold marks the lowest false-positive rate and the highest recall and balanced accuracy.}
\label{tab:fixed_threshold}
\centering
\small
\setlength{\tabcolsep}{3pt}
\begin{tabular*}{\linewidth}{@{\extracolsep{\fill}}lrccc@{}}
\toprule
\multirow{2}{*}{Method} & \multirow{2}{*}{Threshold} & GEN & DetectRL-X & HART \\
\cmidrule(lr){3-3}\cmidrule(lr){4-4}\cmidrule(lr){5-5}
& & \textbf{FPR $\downarrow$ / TPR $\uparrow$ / BA $\uparrow$} & \textbf{FPR $\downarrow$ / TPR $\uparrow$ / BA $\uparrow$} & \textbf{FPR $\downarrow$ / TPR $\uparrow$ / BA $\uparrow$} \\
\midrule
Log-Likelihood & $-1.69295077$ & 5.40 / 9.85 / 52.23 & \textbf{3.87} / 6.88 / 51.51 & 6.00 / 23.25 / 58.63 \\
LRR & $2.24106137$ & 4.45 / 13.10 / 54.33 & 5.68 / 11.25 / 52.78 & \textbf{2.80} / 39.10 / 68.15 \\
TempTest & $-0.0533681143$ & 5.05 / 19.95 / 57.45 & 23.25 / 53.83 / 65.29 & 9.85 / 79.55 / 84.85 \\
NTS & $0.0827531921$ & 4.25 / 32.20 / 63.98 & 24.17 / 57.18 / 66.51 & 7.50 / 84.45 / 88.48 \\
Binoculars & $-0.922963955$ & 3.75 / 22.80 / 59.53 & 30.20 / 57.10 / 63.45 & 6.55 / 86.40 / 89.93 \\
Fast-DetectGPT & $2.04604385$ & 3.90 / 25.45 / 60.78 & 24.92 / 49.15 / 62.12 & 3.65 / 83.00 / 89.68 \\
Lastde++ & $2.10648095$ & \textbf{3.50} / 40.45 / 68.48 & 22.65 / 45.43 / 61.39 & 3.75 / 84.70 / 90.48 \\
CPD (Ours) & $2.05796176$ & 3.75 / \textbf{77.70} / \textbf{86.98} & 6.95 / \textbf{70.35} / \textbf{81.70} & 3.50 / \textbf{91.65} / \textbf{94.08} \\
\bottomrule
\end{tabular*}
\end{table*}

\paragraph{Threshold calibration.}
In the main comparison, each test ROC curve supplies its own operating point for TPR@5\%. For threshold transfer, we calibrate each of the eight detectors on the same 2,000 human and 2,000 AI texts from the GEN development split. For method $m$, we select the threshold with the highest empirical TPR among those that classify at most 5\% of human development texts as AI-generated.
\begin{equation}
\label{eq:fixed_threshold}
\begin{aligned}
\tau_m
&=\underset{\tau\in\mathcal T_m}{\arg\max}\;
\widehat{\mathrm{TPR}}_{\mathrm{dev},m}(\tau)\\
&\quad\text{s.t. }\widehat{\mathrm{FPR}}_{\mathrm{dev},m}(\tau)\le0.05.
\end{aligned}
\end{equation}
The candidate set $\mathcal T_m$ contains every distinct development score for method $m$ and $+\infty$. We evaluate the candidates without interpolation. Ties favor the lower FPR, then the higher threshold. Across detectors, development FPR ranges from 4.75\% to 5.00\%. For CPD, $\tau_{\mathrm{CPD}}\approx2.057962$ gives 4.95\% FPR, 78.85\% TPR, and 86.95\% balanced accuracy.

\paragraph{Cross-benchmark evaluation.}
We apply each frozen threshold to the H vs.\ A test sets
(2,000 texts per class on GEN and HART, 6,000 on DetectRL-X).
We retain the main detector configurations and scoring backbones,
including $k=1$ for CPD. Method $m$ classifies a text $x$ as
AI-generated when $s_m(x)\ge\tau_m$ and as human-written otherwise.
Table~\ref{tab:fixed_threshold} reports TPR, FPR, and balanced
accuracy, $\mathrm{BA}=(\mathrm{TPR}+1-\mathrm{FPR})/2$, where both rates are fractions.

CPD achieves the highest
TPR and BA on all three benchmarks, with BAs of 86.98\%,
81.70\%, and 94.08\% on GEN, DetectRL-X, and HART,
respectively. CPD's FPR stays below the 5\% target on GEN
(3.75\%) and HART (3.50\%) and rises to 6.95\% on DetectRL-X,
where FPR ranges from 3.87\% to 30.20\% across detectors.

\subsection{Single-Model Fast-DetectGPT and Combinations with TCD and TOCSIN}
\label{app:tcd_fast_comparison}

\begin{table*}[!tp]
\caption{Direct detection (H vs.\ A) with Fast-DetectGPT and CPD, including variants that use TCD or TOCSIN. Both metrics are percentages. A dagger marks detectors that use two causal language models. Both TOCSIN variants use BART-base to score reconstruction of the original text after word deletion. Macro Avg. weights the three benchmarks equally. Bold marks the best value in each column.}
\label{tab:tcd_fast_comparison}
\centering
\small
\setlength{\tabcolsep}{2.8pt}
\begin{tabular*}{\linewidth}{@{\extracolsep{\fill}}lcccccccc@{}}
\toprule
\multirow{2}{*}{Method}
& \multicolumn{2}{c}{GEN}
& \multicolumn{2}{c}{DetectRL-X}
& \multicolumn{2}{c}{HART}
& \multicolumn{2}{c}{Macro Avg.} \\
\cmidrule(lr){2-3}
\cmidrule(lr){4-5}
\cmidrule(lr){6-7}
\cmidrule(lr){8-9}
& AUROC & TPR@5\%
& AUROC & TPR@5\%
& AUROC & TPR@5\%
& AUROC & TPR@5\% \\
\midrule
Single-Fast
& 42.79 & 21.30
& 72.72 & 25.30
& 87.71 & 76.60
& 67.74 & 41.07 \\
Fast$^\dagger$
& 47.05 & 26.50
& 69.38 & 15.72
& 90.78 & 84.30
& 69.07 & 42.17 \\
Fast+TCD$^\dagger$
& 50.51 & 26.90
& 68.92 & 20.02
& 93.11 & 85.45
& 70.85 & 44.12 \\
Fast+TOCSIN$^\dagger$
& 47.77 & 20.40
& 68.35 & 18.02
& 89.68 & 80.30
& 68.60 & 39.57 \\
\midrule
CPD
& \textbf{94.18} & \textbf{80.25}
& \textbf{91.90} & \textbf{65.60}
& \textbf{98.31} & 92.95
& \textbf{94.80} & \textbf{79.60} \\
CPD+TCD
& 93.36 & 78.10
& 90.97 & 63.03
& 98.30 & \textbf{93.25}
& 94.21 & 78.13 \\
CPD+TOCSIN
& 89.42 & 55.75
& 90.97 & 59.77
& 97.44 & 90.10
& 92.61 & 68.54 \\
\bottomrule
\end{tabular*}
\end{table*}

Table~\ref{tab:tcd_fast_comparison} compares Single-Fast~\citep{fastdetectgpt}, dual-model Fast-DetectGPT (Fast), and CPD on H vs.\ A, and evaluates TCD~\citep{tcd} and TOCSIN~\citep{tocsin} with each of the latter two detectors. The unweighted CPD scores come from the main run. A second run supplies Single-Fast, Fast, and both TCD variants, and a third supplies the two TOCSIN variants. All seven configurations score the same test documents, with 2,000 texts per class on GEN and HART and 6,000 per class on DetectRL-X. The causal models run in BF16 with a 1,024-token limit and score positions starting at $t=2$. Each retained document has at least eight scored tokens.

Single-Fast uses Falcon-7B for both scoring and reference distributions. Fast and its variants use Falcon-7B-Instruct for scoring and Falcon-7B as the reference. All CPD variants use Falcon-7B with $k=1$ for the contextual contrast. TCD derives token weights from attention entropy in dynamically selected layers of the scoring model, with weighting temperature $\tau_{\mathrm{TCD}}=0.7$. The weights enter the residual sum and their squares enter the reference variance.

TOCSIN rescales the final document score. For each text, we create ten copies by randomly deleting space-delimited words at a nominal rate of 1.5\%. For each copy, BART-base (FP32) computes the mean per-token loss for reconstructing the original text. Averaging these ten losses gives the cohesiveness score $u$. For both Fast+TOCSIN and CPD+TOCSIN, we multiply the base score $s$ by $\exp(u)$ for either sign of $s$, following the Fast branch of the authors' implementation.\footnote{\url{https://github.com/Shixuan-Ma/TOCSIN/blob/main/TOCSIN.py}} The paper's general rule uses $\exp(-u)$ for negative scores~\citep[Eq.~2]{tocsin}.

All three CPD configurations outperform all four Fast configurations on both metrics across the three benchmarks (Table~\ref{tab:tcd_fast_comparison}). TCD weighting raises Fast-DetectGPT's recall at 5\% FPR on every benchmark, taking macro recall from 42.17\% to 44.12\%. For CPD, weighting lowers both metrics on GEN and DetectRL-X. On HART, recall rises from 92.95\% to 93.25\%, while AUROC changes from 98.31\% to 98.30\%. Unweighted CPD has higher macro recall (79.60\% versus 78.13\%) and AUROC (94.80\% versus 94.21\%).

For Fast+TOCSIN on GEN, AUROC rises from 47.05\% to 47.77\% while TPR@5\% falls from 26.50\% to 20.40\%. On DetectRL-X, TPR@5\% rises from 15.72\% to 18.02\% despite a lower AUROC. On HART, both metrics fall. For CPD, rescaling lowers both metrics on every benchmark. GEN shows the largest recall loss, a drop of 24.50 percentage points in TPR@5\%. Macro-averaged TPR@5\% falls from 79.60\% for CPD to 68.54\% for CPD+TOCSIN.

\subsection{HART Source-Discrimination Details}
\label{app:hart_source_results}

The joint comparison of the eight detectors in Section~\ref{sec:Source_hart} excluded one AH text. Grouping the retained scores by draft origin gives 4,000 H/HA negatives and 3,999 A/AH positives (Table~\ref{tab:hart_origin} and Figure~\ref{fig:content-analysis}). The final-expression labels assign AH to the negative class and HA to the positive class, giving 3,999 H/AH negatives and 4,000 A/HA positives. Both H versus AH and HA versus AH have 2,000 negatives and 1,999 positives. HA versus A has 2,000 per class. The separate component ablations use 2,000 texts per class in each of the three pairwise comparisons (Appendix~\ref{app:hart_component_results}).

\subsection{Domain-Wise Source Discrimination on HART}
\label{app:hart_source_domain_results}

\begin{table*}[!tp]
\caption{Draft-origin detection within HART domains on texts of 300--500 Falcon-7B tokens before truncation (H/HA vs.\ A/AH; AUROC / TPR@5\%, in \%). Counts below each domain give the numbers of human-origin negatives and AI-origin positives. Domain Macro Avg. weights the four domains equally. $^\dagger$ denotes a dual-model configuration. Bold marks the highest AUROC and recall in each column.}
\label{tab:hart_source_domain_results}
\centering
\small
\setlength{\tabcolsep}{3pt}
\begin{tabular*}{\linewidth}{@{\extracolsep{\fill}}lccccc@{}}
\toprule
Method & \shortstack{Essays\\(441 / 481)} & \shortstack{arXiv\\(353 / 415)} & \shortstack{Creative Writing\\(378 / 369)} & \shortstack{News\\(414 / 443)} & \shortstack{Domain\\Macro Avg.} \\
\midrule
Log-Likelihood & 64.87 / 34.72 & 50.54 / 20.72 & 59.14 / 30.62 & 44.53 / 3.16 & 54.77 / 22.31 \\
LRR & 57.53 / 32.02 & 59.98 / 30.84 & 58.90 / 29.81 & 56.23 / 23.93 & 58.16 / 29.15 \\
TempTest & 71.50 / 38.05 & 63.35 / 44.10 & 64.19 / 47.70 & 61.72 / 38.83 & 65.19 / 42.17 \\
NTS & 78.72 / 39.50 & 78.38 / 52.29 & 80.73 / 56.37 & 78.18 / 50.34 & 79.00 / 49.62 \\
Binoculars$^\dagger$ & 75.77 / 41.37 & 69.25 / 49.64 & 70.26 / 55.56 & 69.30 / 47.18 & 71.14 / 48.44 \\
Fast-DetectGPT$^\dagger$ & 75.46 / 39.29 & 69.33 / 51.08 & 70.28 / 55.28 & 69.14 / 50.79 & 71.05 / 49.11 \\
Lastde++$^\dagger$ & 79.65 / 40.75 & 77.27 / 53.73 & 77.73 / 56.64 & 73.91 / 48.31 & 77.14 / 49.86 \\
\midrule
\textbf{CPD$_{k=1}$ (Ours)} & \textbf{85.39} / \textbf{51.98} & \textbf{95.45} / \textbf{81.45} & \textbf{92.30} / \textbf{72.63} & \textbf{90.84} / \textbf{73.59} & \textbf{90.99} / \textbf{69.91} \\
\bottomrule
\end{tabular*}
\end{table*}

We use the draft-origin labels from the pooled comparison in Section~\ref{sec:Source_hart} (Table~\ref{tab:hart_origin}) to evaluate H/HA versus A/AH separately in essays, arXiv, creative writing, and news, with A/AH as positives. After initial class balancing, we retain every text measuring 300--500 Falcon-7B tokens before truncation, inclusive. The column headers in Table~\ref{tab:hart_source_domain_results} give the resulting class counts. All eight detectors use their main-experiment settings across domains, with $k=1$ and scoring from $t=2$ for CPD. We compute AUROC and TPR@5\% from each domain's test ROC and average the four unrounded values for each metric with equal weights.

CPD leads on both metrics in all four domains, with macro averages of 90.99\% AUROC and 69.91\% TPR@5\%. Its AUROC ranges from 85.39\% on essays to 95.45\% on arXiv.

\subsection{Source Discrimination by Humanization Type on HART}
\label{app:hart_humanization_types}

\begin{table*}[!tp]
\caption{HART source discrimination by humanization type (H vs.\ AH). Cells give AUROC / TPR@5\% in percent for human drafts versus humanized AI drafts. Counts below each heading give the numbers of human and humanized AI texts retained after length filtering. The macro average weights the three types equally. $^\dagger$ marks dual-model detectors. Bold marks the highest value of each metric.}
\label{tab:hart_humanization_types}
\centering
\small
\setlength{\tabcolsep}{4pt}
\begin{tabular*}{\linewidth}{@{\extracolsep{\fill}}lcccc@{}}
\toprule
Method & \shortstack{Model\\(578 / 614)} & \shortstack{Tool\\(155 / 145)} & \shortstack{Human\\(91 / 101)} & \shortstack{Type\\Macro Avg.} \\
\midrule
Log-Likelihood & 35.23 / 0.98 & 25.11 / 1.38 & 57.37 / 0.99 & 39.24 / 1.12 \\
LRR & 30.19 / 1.79 & 29.37 / 2.76 & 53.23 / 9.90 & 37.60 / 4.82 \\
TempTest & 49.97 / 15.80 & 35.41 / 8.97 & 81.41 / 53.47 & 55.60 / 26.08 \\
NTS & 76.50 / 43.16 & 66.35 / 24.83 & 86.60 / 60.40 & 76.48 / 42.79 \\
Binoculars$^\dagger$ & 60.97 / 38.44 & 49.37 / 17.24 & \textbf{92.32} / 78.22 & 67.55 / 44.63 \\
Fast-DetectGPT$^\dagger$ & 60.96 / 39.41 & 49.73 / 21.38 & 92.02 / \textbf{80.20} & 67.57 / 47.00 \\
Lastde++$^\dagger$ & 72.29 / 40.23 & 60.81 / 12.41 & 91.86 / 72.28 & 74.99 / 41.64 \\
\midrule
\textbf{CPD$_{k=1}$ (Ours)} & \textbf{95.49} / \textbf{86.97} & \textbf{80.37} / \textbf{46.21} & 92.18 / 69.31 & \textbf{89.35} / \textbf{67.49} \\
\bottomrule
\end{tabular*}
\end{table*}

We compare original human texts (H) with humanized AI-origin texts (AH), grouping the latter by who or what rewrote the AI draft. The Model group includes rewrites by GPT-3.5-Turbo, GPT-4o, Gemini-1.5-Pro, and Claude-3.5-Sonnet. The Tool and Human groups contain rewrites by a humanization tool and by human editors, respectively. We assign each H text to the group of its corresponding AH text using the HART domain and source ID.

Following Appendix~\ref{app:hart_source_domain_results}, we filter each H and AH document individually, retaining texts of 300--500 Falcon-7B tokens (inclusive, before truncation). Table~\ref{tab:hart_humanization_types} gives the retained counts by group. For the eight detector configurations in Table~\ref{tab:hart_origin}, we compute AUROC and TPR@5\% from each group's test ROC with AH as the positive class.

At 5\% FPR, CPD detects 86.97\% of model rewrites and 46.21\% of tool rewrites (Table~\ref{tab:hart_humanization_types}). On human rewrites, CPD and Binoculars have similar AUROCs (92.18\% and 92.32\%), while Fast-DetectGPT detects the most rewrites at 5\% FPR (80.20\%, compared with 69.31\% for CPD).

\subsection{Source-Matched AI-mediated Rewriting}
\label{app:ai_rewriting_results}

\begin{table*}[!tp]
\caption{Draft-origin detection on GEN before and after AI editing of the same human drafts (AUROC and TPR@5\%, in \%). The negatives are 936 human drafts or their revisions within the indicated edit-ratio range. The same 936 directly generated AI texts serve as positives throughout. Bold marks the best detector for each metric on the original drafts and within each of the three edit-ratio ranges.}
\label{tab:gen_edit_robustness}
\centering
\small
\setlength{\tabcolsep}{3pt}
\begin{tabular*}{\linewidth}{@{\extracolsep{\fill}}lcccccccc@{}}
\toprule
\multirow{2}{*}{Method} & \multicolumn{2}{c}{Original (0\%)} & \multicolumn{2}{c}{20--40\%} & \multicolumn{2}{c}{40--60\%} & \multicolumn{2}{c}{60--80\%} \\
\cmidrule(lr){2-3}\cmidrule(lr){4-5}\cmidrule(lr){6-7}\cmidrule(lr){8-9}
& AUROC & TPR@5\% & AUROC & TPR@5\% & AUROC & TPR@5\% & AUROC & TPR@5\% \\
\midrule
LogLikelihood
& 47.09 & 10.04
& 43.04 & 8.97
& 38.76 & 6.41
& 34.66 & 4.27 \\
LRR
& 51.61 & 17.84
& 52.78 & 17.84
& 50.58 & 13.68
& 47.72 & 8.97 \\
TempTest
& 44.76 & 19.34
& 35.86 & 13.35
& 30.34 & 7.48
& 24.63 & 3.63 \\
NTS
& 66.67 & 33.23
& 55.44 & 21.90
& 46.24 & 13.35
& 37.53 & 6.84 \\
Binoculars
& 45.94 & 23.29
& 33.20 & 14.74
& 27.52 & 7.91
& 22.03 & 3.10 \\
Fast-DetectGPT
& 46.79 & 25.64
& 34.36 & 16.45
& 29.20 & 9.51
& 24.72 & 4.91 \\
Lastde++
& 68.96 & 43.59
& 55.35 & 26.07
& 48.54 & 19.02
& 41.80 & 9.08 \\
\midrule
CPD$_{k=1}$ (Ours)
& \textbf{93.51} & \textbf{79.06}
& \textbf{88.13} & \textbf{69.44}
& \textbf{82.33} & \textbf{58.44}
& \textbf{77.71} & \textbf{52.35} \\
\bottomrule
\end{tabular*}
\end{table*}

\paragraph{GEN: editing intensity on matched human drafts.}
We select 936 human drafts with one AI-edited version in each of the 20--40\%, 40--60\%, and 60--80\% edit-ratio ranges, yielding 2,808 revisions. For all eight detectors, the negatives are the original drafts in the H vs.\ A condition and the corresponding 936 revisions in each HA vs.\ A condition (Table~\ref{tab:gen_edit_robustness} and Figure~\ref{fig:gen_detectrlx_edit}(a)). The human drafts come from a separate prompt pool, while all four conditions share 936 directly AI-generated positives drawn from the 2,000-per-class direct-detection test set in Table~\ref{tab:main_results}.

The original human drafts, the three edited sets, and the AI references have the same domain composition, with 196 arXiv, 188 Reddit, 196 story-generation, 179 wikiHow, and 177 Wikipedia texts per condition. Llama-3.3-70B, GPT-OSS-120B, and Qwen2.5-72B each contributed 312 revisions within each edit-ratio range. For 825 of the 936 drafts, the three revisions come from at least two editing models.

\paragraph{DetectRL-X: editing operations on matched AI drafts.}
Using source-pair identifiers, we match each of the 6,000 original AI drafts to its condensed, expanded, and polished versions. All four conditions share the same 6,000 human controls. Each condition contains 1,000 AI texts from each of six domains and 1,500 from each of four generators.

Table~\ref{tab:aa_robustness} reports AUROC and TPR@5\% for each DetectRL-X condition in Figure~\ref{fig:gen_detectrlx_edit}. For the Overall column, we draw 6,000 AI variants from the pool of individually condensed, expanded, and polished texts and compute one ROC curve against the 6,000 human controls (Appendix~\ref{sec:data_eval_settings}).

\begin{table*}[!tp]
\caption{
Detection of AI drafts after rewriting on DetectRL-X (AUROC and TPR@5\%, in \%). Each condition uses the same 6,000 human controls. Original is unmodified, and Overall pools the three refinement types. Bold marks the best result per metric and condition.
}
\label{tab:aa_robustness}
\centering
\small
\setlength{\tabcolsep}{3pt}
\begin{tabular*}{\linewidth}{@{\extracolsep{\fill}}lcccccccccc@{}}
\toprule
\multirow{2}{*}{Method}
& \multicolumn{2}{c}{Original}
& \multicolumn{2}{c}{Condensing}
& \multicolumn{2}{c}{Expanding}
& \multicolumn{2}{c}{Polishing}
& \multicolumn{2}{c}{Overall} \\
\cmidrule(lr){2-3}
\cmidrule(lr){4-5}
\cmidrule(lr){6-7}
\cmidrule(lr){8-9}
\cmidrule(lr){10-11}
& AUROC & TPR@5\%
& AUROC & TPR@5\%
& AUROC & TPR@5\%
& AUROC & TPR@5\%
& AUROC & TPR@5\% \\
\midrule
LogLikelihood
& 55.52 & 8.72
& 36.39 & 0.85
& 65.15 & 2.57
& 31.54 & 0.62
& 44.25 & 1.30 \\

LRR
& 53.33 & 10.57
& 32.61 & 0.68
& 62.74 & 1.80
& 25.26 & 0.73
& 40.06 & 1.08 \\

TempTest
& 72.74 & 21.80
& 45.75 & 1.55
& 86.23 & 23.70
& 40.24 & 2.15
& 57.34 & 9.03 \\

NTS
& 72.73 & 19.92
& 48.57 & 1.30
& 84.98 & 16.13
& 44.67 & 2.03
& 59.28 & 6.47 \\

Binoculars
& 68.42 & 9.22
& 45.49 & 0.02
& 77.59 & 0.27
& 39.57 & 0.42
& 54.08 & 0.23 \\

Fast-DetectGPT
& 69.39 & 15.70
& 43.39 & 0.63
& 88.23 & 31.50
& 40.88 & 3.43
& 57.28 & 11.83 \\

Lastde++
& 68.40 & 11.85
& 42.98 & 0.23
& 89.42 & 29.10
& 43.49 & 2.13
& 58.47 & 10.50 \\

\midrule
CPD$_{k=1}$ (Ours)
& \textbf{91.90} & \textbf{65.60}
& \textbf{88.27} & \textbf{53.63}
& \textbf{99.98} & \textbf{99.90}
& \textbf{84.59} & \textbf{44.93}
& \textbf{90.78} & \textbf{65.57} \\
\bottomrule
\end{tabular*}
\end{table*}

\subsection{Length-Controlled Evaluation}
\label{sec:length_control}

\begin{table}[!htbp]
\caption{Macro AUROC / TPR@5\% (\%) on matched prefixes of GEN, DetectRL-X, and HART texts with at least 256 tokens each.}
\label{tab:length_summary}
\centering
\small
\setlength{\tabcolsep}{2pt}
\renewcommand{\arraystretch}{0.95}
\begin{tabular*}{\linewidth}{@{\extracolsep{\fill}}lccc@{}}
\toprule
Method & 64 tokens & 128 tokens & 256 tokens \\
\midrule
Loglikelihood & 61.43 / 4.75 & 63.62 / 6.17 & 63.88 / 9.64 \\
LRR & 58.76 / 8.42 & 64.10 / 14.73 & 66.93 / 22.37 \\
TempTest & 68.01 / 22.48 & 73.47 / 34.48 & 73.53 / 42.92 \\
NTS & 64.05 / 20.36 & 73.97 / 36.11 & 77.86 / 46.46 \\
Binoculars & 70.11 / 30.65 & 74.87 / 44.87 & 73.19 / 44.96 \\
Fast-DetectGPT & 70.05 / 30.64 & 74.93 / 45.10 & 73.40 / 44.53 \\
Lastde++ & 70.40 / 27.40 & 77.81 / 43.18 & 79.21 / 46.84 \\
\midrule
\textbf{CPD$_{k=1}$ (Ours)} & \textbf{81.49 / 39.90} & \textbf{90.26 / 62.78} & \textbf{95.10 / 80.05} \\
\bottomrule
\end{tabular*}
\end{table}

\begin{table*}[!tp]
\caption{Length-controlled detection on GEN, DetectRL-X, and HART (H vs.\ A; AUROC / TPR@5\%, in \%). We retain texts measuring at least 256 tokens before truncation. Each detector scores the first 64, 128, or 256 tokens of the same texts. At each prefix length, Macro Avg. gives the three benchmarks equal weight. Bold marks the highest AUROC and TPR@5\% within each column at each length.}
\label{tab:length_control}
\centering
\small
\setlength{\tabcolsep}{3pt}
\begin{tabular*}{\linewidth}{@{\extracolsep{\fill}}clcccc@{}}
\toprule
Length & Method & GEN & DetectRL-X & HART & Macro Avg. \\
\midrule
\multirow{8}{*}{64} & LogLikelihood & 57.34 / 7.36 & 61.55 / 5.80 & 65.41 / 1.08 & 61.43 / 4.75 \\
 & LRR & 58.67 / 9.23 & 54.87 / 5.68 & 62.74 / 10.34 & 58.76 / 8.42 \\
 & TempTest & 57.91 / 15.22 & 64.33 / 14.90 & 81.78 / 37.31 & 68.01 / 22.48 \\
 & NTS & 58.94 / 15.85 & 56.13 / 10.78 & 77.09 / 34.45 & 64.05 / 20.36 \\
 & Binoculars & 61.22 / 25.76 & 60.87 / 12.81 & 88.25 / 53.36 & 70.11 / 30.65 \\
 & Fast-DetectGPT & 61.00 / 24.95 & 60.82 / 12.34 & 88.33 / \textbf{54.63} & 70.05 / 30.64 \\
 & Lastde++ & 69.33 / 26.82 & 58.57 / 9.49 & 83.28 / 45.88 & 70.40 / 27.40 \\
 & CPD$_{k=1}$ (Ours) & \textbf{79.73} / \textbf{37.49} & \textbf{76.34} / \textbf{28.90} & \textbf{88.42} / 53.30 & \textbf{81.49} / \textbf{39.90} \\
\midrule
\multirow{8}{*}{128} & LogLikelihood & 59.09 / 7.92 & 59.66 / 7.24 & 72.11 / 3.36 & 63.62 / 6.17 \\
 & LRR & 63.30 / 14.78 & 56.01 / 8.10 & 73.01 / 21.32 & 64.10 / 14.73 \\
 & TempTest & 58.23 / 21.71 & 71.32 / 22.59 & 90.85 / 59.14 & 73.47 / 34.48 \\
 & NTS & 63.94 / 23.27 & 66.79 / 20.46 & 91.19 / 64.59 & 73.97 / 36.11 \\
 & Binoculars & 60.24 / 30.69 & 68.76 / 20.93 & 95.60 / 82.99 & 74.87 / 44.87 \\
 & Fast-DetectGPT & 60.10 / 29.57 & 69.03 / 21.85 & 95.67 / \textbf{83.88} & 74.93 / 45.10 \\
 & Lastde++ & 72.48 / 34.81 & 66.69 / 18.26 & 94.26 / 76.46 & 77.81 / 43.18 \\
 & CPD$_{k=1}$ (Ours) & \textbf{89.69} / \textbf{60.76} & \textbf{84.61} / \textbf{43.96} & \textbf{96.47} / 83.63 & \textbf{90.26} / \textbf{62.78} \\
\midrule
\multirow{8}{*}{256} & LogLikelihood & 56.55 / 9.42 & 57.75 / 8.53 & 77.33 / 10.98 & 63.88 / 9.64 \\
 & LRR & 64.09 / 16.47 & 54.42 / 9.65 & 82.27 / 40.99 & 66.93 / 22.37 \\
 & TempTest & 53.94 / 23.83 & 74.03 / 24.92 & 92.62 / 80.01 & 73.53 / 42.92 \\
 & NTS & 65.65 / 29.26 & 72.76 / 23.90 & 95.18 / 86.23 & 77.86 / 46.46 \\
 & Binoculars & 54.00 / 30.19 & 69.99 / 13.63 & 95.56 / 91.05 & 73.19 / 44.96 \\
 & Fast-DetectGPT & 53.93 / 29.07 & 70.74 / 13.79 & 95.54 / 90.74 & 73.40 / 44.53 \\
 & Lastde++ & 71.73 / 38.80 & 69.32 / 10.04 & 96.58 / 91.69 & 79.21 / 46.84 \\
 & CPD$_{k=1}$ (Ours) & \textbf{95.24} / \textbf{82.16} & \textbf{91.07} / \textbf{61.92} & \textbf{98.99} / \textbf{96.07} & \textbf{95.10} / \textbf{80.05} \\
\bottomrule
\end{tabular*}
\end{table*}

We evaluate all eight detectors on H vs.\ A across GEN, DetectRL-X, and HART. Texts with fewer than 256 tokens before truncation are excluded, and each retained text is truncated to 64, 128, or 256 tokens. Every method scores the same texts at all three lengths. After filtering, GEN contains 1,796 human and 1,603 AI texts, DetectRL-X contains 5,376 human and 4,879 AI texts, and HART contains 1,615 human and 1,576 AI texts. We average the three dataset results before rounding, giving each dataset equal weight.

CPD's macro AUROC and TPR@5\% increase with prefix length. CPD has the highest AUROC at every prefix length on all three benchmarks (Table~\ref{tab:length_control}). Fast-DetectGPT has higher TPR@5\% on HART at 64 and 128 tokens. At 256 tokens, CPD leads on both metrics across benchmarks.

\subsection{Character-Level Perturbations}
\label{app:character_robustness_results}

\begin{figure*}[!tp]
    \centering
    \includegraphics[width=0.9\linewidth]
    {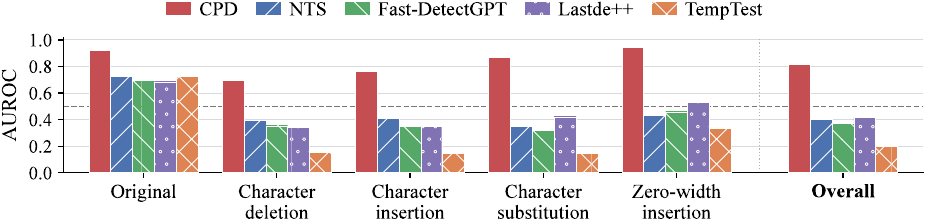}
    \caption{
Detection under character perturbations on DetectRL-X (AUROC, 0--1). The same 6,000 human texts serve as controls in all conditions. Original uses the unmodified AI drafts. Overall uses 6,000 perturbed AI texts sampled across all four character operations.
}
    \label{fig:char_robustness}
\end{figure*}

CPD's AUROC ranges from 69.37\% after character deletion to 94.08\% after zero-width insertion (Figure~\ref{fig:char_robustness}). Pooling 6,000 sampled AI variants across the four operations against the human controls gives 81.38\% AUROC (Appendix~\ref{sec:dataset_detectrlx_robustness}).

\subsection{Cross-Lingual Generalization}
\label{app:cross_lingual_results}

\begin{table*}[!tp]
\caption{Detection on eight DetectRL-X language test sets, each with 6,000 human texts and 6,000 AI texts (all values in \%). The upper panel reports AUROC for all eight detectors, with the best value in each column in bold. The lower panel gives CPD's FPR and TPR in each language at the threshold calibrated to a 5\% FPR target on GEN development data. Macro Avg. weights languages equally.}
\label{tab:cross_lingual}
\centering
\small
\setlength{\tabcolsep}{2.5pt}
\begin{tabular*}{\linewidth}{@{\extracolsep{\fill}}lccccccccc@{}}
\toprule
Method & Arabic & Chinese & English & French & German & Portuguese & Russian & Spanish & Macro Avg. \\
\midrule
LogLikelihood & 52.80 & 74.48 & 55.52 & 74.47 & 79.90 & 78.65 & 64.03 & 79.93 & 69.97 \\
LRR & 49.22 & 67.47 & 53.33 & 65.90 & 65.42 & 65.90 & 56.55 & 64.45 & 61.03 \\
TempTest & 63.05 & 84.11 & 72.74 & 90.39 & 85.16 & \textbf{88.03} & \textbf{71.09} & 85.24 & 79.98 \\
NTS & 56.61 & 80.41 & 72.73 & 87.86 & 79.99 & 84.94 & 66.60 & 82.12 & 76.41 \\
Binoculars & 35.85 & 61.43 & 68.42 & 84.46 & 68.65 & 73.73 & 44.65 & 74.63 & 63.98 \\
Fast-DetectGPT & 39.67 & 61.73 & 69.39 & 83.97 & 68.19 & 73.48 & 46.71 & 74.33 & 64.68 \\
Lastde++ & 43.91 & 62.78 & 68.40 & 77.98 & 58.36 & 66.38 & 45.55 & 67.39 & 61.34 \\
\midrule
CPD$_{k=1}$ (Ours) & \textbf{64.81} & \textbf{84.61} & \textbf{91.90} & \textbf{93.86} & \textbf{85.55} & 84.89 & 67.94 & \textbf{88.80} & \textbf{82.80} \\
\midrule
\multicolumn{10}{l}{\textit{CPD at a fixed threshold (target FPR of 5\% on GEN development data)}} \\
FPR & 11.27 & 4.60 & 6.95 & 7.95 & 10.72 & 6.23 & 7.17 & 11.82 & 8.34 \\
TPR & 22.55 & 43.93 & 70.35 & 79.10 & 61.07 & 49.97 & 20.95 & 70.07 & 52.25 \\
\bottomrule
\end{tabular*}
\end{table*}

Table~\ref{tab:cross_lingual} reports AUROC by language for all eight detectors in their default configurations. Each test set contains 6,000 human and 6,000 AI texts. We compute equally weighted macro averages over all eight languages and over the seven non-English languages, using unrounded results in both cases. In every language, CPD classifies a text as AI-origin if its score reaches or exceeds the full-precision GEN development threshold ($\tau\approx 2.057962$, Appendix~\ref{app:threshold_transfer}). The English test-set results reproduce the DetectRL-X AUROC in Table~\ref{tab:main_results} and the two fixed-threshold rates in Table~\ref{tab:fixed_threshold}.

Arabic and Russian have CPD's lowest AUROCs. The frozen threshold also detects the smallest shares of AI texts in these two languages, 22.55\% and 20.95\%, respectively.

\subsection{Cross-Backbone Generalization}
\label{app:cross_backbone_results}

\begin{table*}[!tp]
\caption{
Single-model detection on the GEN test set by scoring backbone (AUROC / TPR@5\%, in \%). Single-Fast uses the same backbone for the scoring and reference distributions in Fast-DetectGPT. Bold marks the best value for each backbone and metric.
}
\label{tab:cross_backbone_detailed}
\centering
\small
\setlength{\tabcolsep}{4pt}
\begin{tabular*}{\linewidth}{@{\extracolsep{\fill}}lccccc@{}}
\toprule
\multirow{2}{*}{Method}
& \multicolumn{5}{c}{Backbone Models} \\
\cmidrule(lr){2-6}
& Falcon-7B
& Falcon-7B-Instruct
& GPT-Neo-2.7B
& Llama-2-7B
& Llama-3.1-8B \\
\midrule
Loglikelihood
& 43.04 / 8.65
& 43.04 / 11.50
& 37.35 / 3.60
& 53.17 / 9.85
& 53.94 / 10.40 \\

LRR
& 47.94 / 13.65
& 50.41 / 15.25
& 40.08 / 6.80
& 51.68 / 14.45
& 61.38 / 25.55 \\

TempTest
& 43.91 / 19.90
& 27.30 / 8.20
& 29.34 / 12.95
& 51.45 / 17.65
& 54.08 / 18.70 \\

NTS
& 67.75 / 33.55
& 57.88 / 31.30
& 76.48 / 49.35
& 71.13 / 33.75
& 66.84 / 27.75 \\

Single-Fast
& 42.79 / 21.30
& 21.33 / 8.15
& 26.34 / 13.65
& 42.42 / 15.90
& 46.49 / 18.55 \\

\midrule
CPD (Ours)
& \textbf{94.18 / 80.25}
& \textbf{80.41 / 59.15}
& \textbf{92.08 / 74.35}
& \textbf{91.96 / 74.05}
& \textbf{87.02 / 62.25} \\
\bottomrule
\end{tabular*}
\end{table*}

Table~\ref{tab:cross_backbone_detailed} compares six detectors on the same 2,000 human-written and 2,000 AI-generated texts from the English GEN test set. Each column uses one Falcon~\citep{falcon}, GPT-Neo~\citep{gpt-neo}, or Llama~\citep{llama2,llama3} backbone for all six methods. Loglikelihood scores the observed tokens under the full prefix. Single-Fast uses the same full-prefix distribution for scoring and for its reference moments. CPD uses that backbone for both the full prefix and the one-token prefix ($k=1$). We keep each detector's score direction fixed across backbones, as in Section~\ref{sec:exp_setup}. The Falcon-7B column reproduces the GEN results in Table~\ref{tab:main_results}.

NTS ranks second on both metrics with every backbone. Its closest result to CPD is on GPT-Neo-2.7B, where CPD leads by 15.60 percentage points in AUROC and 25.00 in TPR@5\%. Loglikelihood and Single-Fast both attain their highest AUROC on Llama-3.1-8B (53.94\% and 46.49\%, respectively). With that same scorer, CPD reaches 87.02\% AUROC and detects 62.25\% of AI texts at 5\% FPR, compared with 18.55\% for Single-Fast.

\paragraph{Effect of instruction tuning.}
On the same texts with $k=1$, switching from Falcon-7B to Falcon-7B-Instruct narrows the gap between mean CPD scores for AI and human drafts from 4.50 to 3.13 (computed before rounding). Recall at 5\% FPR falls from 80.25\% to 59.15\% (Table~\ref{tab:cross_backbone_detailed}). On a separate writing benchmark, \citet[Table~15]{nts} also report lower NTS AUROC with instruction-tuned Falcon.

\subsection{Computational Efficiency}
\label{app:efficiency_results}

\newsavebox{\efficiencytablebox}
\begin{lrbox}{\efficiencytablebox}
\begin{minipage}[c]{0.49\textwidth}
\captionsetup{type=table}
\caption{Computational cost on 200 GEN texts using one NVIDIA RTX A6000. Peak allocated GPU memory includes model weights. Timing begins after model and data loading and two warm-up texts.}
\label{tab:method_cost}
\centering
\small
\begin{tabular*}{\columnwidth}{@{\extracolsep{\fill}}lcc@{}}
\toprule
Method & Memory (GB) & Time (s/text) \\
\midrule
Loglikelihood & 14.21 & 0.113 \\
LRR & 14.21 & 0.117 \\
TempTest & 14.21 & 0.119 \\
NTS & 14.21 & 0.122 \\
Binoculars & 28.19 & 0.234 \\
Fast-DetectGPT & 28.19 & 0.233 \\
Lastde++ & 28.19 & 0.306 \\
\midrule
CPD$_{k=1}$ (Ours) & 15.80 & 0.176 \\
\bottomrule
\end{tabular*}
\end{minipage}
\end{lrbox}

\begin{figure*}[!tp]
\centering
\begin{minipage}[c]{0.48\textwidth}
    \centering
    \includegraphics[width=\linewidth]
    {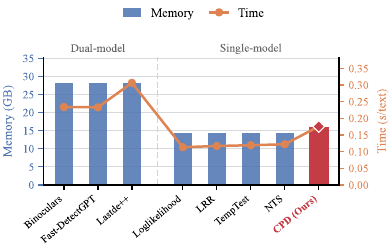}
    \caption{
Peak allocated GPU memory (bars, GB) and mean time per text (line, seconds) on 200 GEN texts with one RTX A6000.
}
    \label{fig:cost}
\end{minipage}\hfill
\usebox{\efficiencytablebox}
\end{figure*}

\begin{table*}[!tp]
\caption{Direct detection on GEN for eight generators. Each prompt pairs a human text with an AI output. Each column contains 250 prompt pairs, except Gemma 3-27B (251) and GPT-OSS-20B (249). Bold marks the highest value for each generator and metric.}
\label{tab:gen_generator_results}
\centering
\small
\setlength{\tabcolsep}{3pt}
\begin{tabular*}{\linewidth}{@{\extracolsep{\fill}}l*{8}{r}@{}}
\toprule
 & \multicolumn{2}{c}{Gemma 3} & \multicolumn{2}{c}{Qwen2.5} & \multicolumn{2}{c}{GPT-OSS} & \multicolumn{2}{c}{Llama} \\
\cmidrule(lr){2-3}\cmidrule(lr){4-5}\cmidrule(lr){6-7}\cmidrule(lr){8-9}
Detector & 12B & 27B & 7B & 72B & 20B & 120B & 3.1-8B & 3.3-70B \\
\midrule
\multicolumn{9}{c}{\textit{AUROC (\%)}} \\
Loglikelihood & 38.02 & 41.32 & 29.96 & 64.16 & 2.84 & 78.34 & 33.95 & 53.24 \\
LRR & 41.93 & 43.31 & 48.32 & 67.43 & 3.00 & 78.48 & 44.56 & 54.03 \\
TempTest & 32.52 & 37.79 & 28.54 & 65.44 & 6.22 & 89.63 & 35.15 & 51.88 \\
NTS & 50.00 & 53.15 & 64.78 & 74.01 & \textbf{85.36} & 89.80 & 53.23 & 71.31 \\
Binoculars & 29.35 & 27.50 & 28.71 & 67.68 & 24.05 & 86.07 & 48.71 & 57.81 \\
Fast-DetectGPT & 29.79 & 29.16 & 30.09 & 67.53 & 26.05 & 86.22 & 49.15 & 58.00 \\
Lastde++ & 59.30 & 49.96 & 67.64 & 78.23 & 41.23 & 86.11 & 87.48 & 77.88 \\
\addlinespace[2pt]
CPD$_{k=1}$ & \textbf{96.15} & \textbf{97.01} & \textbf{97.70} & \textbf{94.98} & 78.42 & \textbf{96.44} & \textbf{97.53} & \textbf{96.83} \\
\midrule
\multicolumn{9}{c}{\textit{TPR at 5\% FPR (\%)}} \\
Loglikelihood & 2.00 & 1.99 & 1.20 & 24.80 & 0.00 & 36.40 & 0.80 & 19.60 \\
LRR & 4.80 & 3.59 & 3.60 & 30.00 & 0.00 & 46.40 & 2.80 & 17.20 \\
TempTest & 0.00 & 1.20 & 4.40 & 47.60 & 0.00 & 76.40 & 8.40 & 25.60 \\
NTS & 10.40 & 7.57 & 28.40 & 44.40 & \textbf{60.24} & 67.20 & 20.40 & 30.40 \\
Binoculars & 1.60 & 2.39 & 3.20 & 50.80 & 4.42 & 70.00 & 21.60 & 40.80 \\
Fast-DetectGPT & 2.40 & 4.78 & 8.40 & 52.00 & 4.42 & 72.40 & 24.80 & 42.00 \\
Lastde++ & 17.20 & 19.52 & 41.20 & 59.20 & 11.24 & 73.20 & 67.20 & 55.60 \\
\addlinespace[2pt]
CPD$_{k=1}$ & \textbf{83.20} & \textbf{85.26} & \textbf{90.40} & \textbf{82.40} & 37.35 & \textbf{86.40} & \textbf{91.20} & \textbf{87.20} \\
\bottomrule
\end{tabular*}
\end{table*}

Figure~\ref{fig:cost} and Table~\ref{tab:method_cost} report mean processing time per text and peak allocated GPU memory on 200 GEN texts using one NVIDIA RTX A6000. CPD reuses predictions for identical one-token restricted contexts within each text. Peak memory includes model weights. Appendix~\ref{sec:data_eval_settings} gives the sampling, timing, and warm-up protocol.

\paragraph{Computational cost.}
CPD uses 15.80\,GB of peak allocated GPU memory and averages 0.176 seconds per text. The dual-model detectors each use 28.19\,GB and take 0.233--0.306 seconds per text. Relative to Fast-DetectGPT, the fastest dual-model baseline, CPD reduces time per text by 24.6\% (a $1.33\times$ speedup) and peak memory by 43.95\%.

Loglikelihood, LRR, TempTest, and NTS use 14.21\,GB and take 0.113--0.122 seconds on average to score one GEN text.

\subsection{Generator-wise Performance Analysis on GEN}
\label{app:gen_generator_results}

Each of the 2,000 prompts in the GEN direct-detection test set contributes one human text and one AI text. We group these pairs by the model that generated the AI text. CPD uses Falcon-7B with $k=1$ across all eight subsets, and each baseline retains its configuration from Section~\ref{sec:exp_setup}. We keep each detector's score direction fixed and compute its ROC curve separately for each generator (Table~\ref{tab:gen_generator_results}). The GEN column in Table~\ref{tab:main_results} evaluates the same texts together.

CPD leads on both metrics for seven of the eight generators. Among those seven generators, its lowest AUROC (94.98\%) and TPR@5\% (82.40\%) occur on Qwen2.5-72B. CPD reaches its highest AUROC of 97.70\% on Qwen2.5-7B and its highest TPR@5\% of 91.20\% on Llama 3.1-8B.

On GPT-OSS-20B, NTS leads with 85.36\% AUROC and 60.24\% TPR@5\%, compared with CPD's 78.42\% and 37.35\%. On GPT-OSS-120B, CPD detects 86.40\% of outputs at 5\% FPR, 49.05 percentage points higher than on GPT-OSS-20B. The difference between mean CPD scores for generated and prompt-matched human texts rises from 1.52 on the 20B subset to 3.43 on the 120B subset.

\FloatBarrier

\end{document}